\documentclass{article}%
\usepackage{amsmath}
\usepackage{amsfonts}
\usepackage{amssymb}
\usepackage{graphicx}%
\providecommand{\U}[1]{\protect\rule{.1in}{.1in}}

\begin{document}

\title{Attention-like feature explanation for tabular data}
\author{Andrei V. Konstantinov and Lev V. Utkin\\Peter the Great St.Petersburg Polytechnic University\\St.Petersburg, Russia\\e-mail: andrue.konst@gmail.com, lev.utkin@gmail.com}
\date{}
\maketitle

\begin{abstract}
A new method for local and global explanation of the machine learning
black-box model predictions by tabular data is proposed. It is implemented as
a system called AFEX (Attention-like Feature EXplanation) and consisting of
two main parts. The first part is a set of the one-feature neural subnetworks
which aim to get a specific representation for every feature in the form of a
basis of shape functions. The subnetworks use shortcut connections with
trainable parameters to improve the network performance. The second part of
AFEX produces shape functions of features as the weighted sum of the basis
shape functions where weights are computed by using an attention-like
mechanism. AFEX identifies pairwise interactions between features based on
pairwise multiplications of shape functions corresponding to different
features. A modification of AFEX with incorporating an additional surrogate
model which approximates the black-box model is proposed. AFEX is trained
end-to-end on a whole dataset only once such that it does not require to train
neural networks again in the explanation stage. Numerical experiments with
synthetic and real data illustrate AFEX.


\end{abstract}

\section{Introduction}

An unprecedented growth of contribution of machine learning methods and models
into solving various significant applied problems leads to requirements to
explain the corresponding model predictions because most powerful machine
learning models are very complex, for example, deep neural networks, their
functioning is unclear due to the black-box nature of the models. The
explanation problem is especially important in such applications as medicine,
safety maintenance, finance, etc. A machine learning model user has to
understand why the model predicts a certain decision, for example, a certain
disease of a patient. This understanding may help to improve the user-model
interaction and to justify the actions following the obtained prediction, for
example, to choose the best treatment in accordance with the stated diagnosis
\cite{Holzinger-etal-2019}. The importance of explaining predictions provided
by black-box models impelled to develop the corresponding explanation methods
and models
\cite{Arya-etal-2019,Belle-Papantonis-2020,Guidotti-2019,Liang-etal-2021,Molnar-2019,Murdoch-etal-2019,Xie-Ras-etal-2020,Zablocki-etal-21,Zhang-Tino-etal-2020}%
.

Depending on the instance set to be explained, there are local and global
methods, which explain the black-box model predictions locally around a test
instance and around a set of instance, for example, around the whole dataset,
respectively. We focus on both the groups of interpretation methods, including
local as well as global interpretations. Among the local explanation methods,
the well-known method is LIME (Local Interpretable Model-Agnostic Explanation)
\cite{Ribeiro-etal-2016}, which belongs to methods based on constructing a
surrogate model \cite{Poyiadzi-etal-21} approximating the black-box model at a
point. The surrogate model in LIME is the linear regression whose coefficients
can be interpreted as the feature importance measures.

Another set of the explanation methods is based on applying Generalized
Additive Models (GAMs) \cite{Hastie-Tibshirani-1990}. GAMs in the framework of
explanation methods can be regarded as surrogate models which are represented
as a sum of influence functions of features, i.e. the GAM outcome is a linear
combination of some functions of features. Several explanation models based on
GAMs have been proposed, including the well-known Explainable Boosting Machine
(EBM) \cite{Nori-etal-19}, the Neural Additive Model (NAM)
\cite{Agarwal-etal-20}, GAMI-Net \cite{Yang-Zhang-Sudjianto-20}, the Adaptive
Explainable Neural Networks \cite{Chen-Vaughan-etal-20}. The main peculiarity
of the aforementioned surrogate models is that the influence or shape
functions from GAM are obtained by training neural networks (in NAM or
GAMI-Net) or by training the functions iteratively (EBM).

There exist many other explanation models, for example, the SHapley Additive
exPlanations (SHAP) \cite{Lundberg-Lee-2017,Strumbel-Kononenko-2010} and its
various modifications. However, we draw attention mainly on the surrogate
models which use GAMs as a basis for explanation because the proposed
explanation method and its modification are also based on GAM to some extent.
We propose a method which uses the attention-like approach. Therefore, we
briefly consider the attention mechanism and its use in solving the
explanation problems.

One of the important and efficient approaches for improving machine learning
models is the attention mechanism which can be regarded as an extension of the
Nadaraya-Waston kernel regression \cite{Nadaraya-1964,Watson-1964}. It
originally stems from a property of the human perception to selectively
concentrate on an important part of information and to ignore other
information \cite{Niu-Zhong-Yu-21}. Most applications of the attention
mechanism focus on the computer vision area, including emotion detection,
image-based analysis, visual question answering, etc., and the natural
language processing (NLP), including text classification, translation, etc.
Comprehensive surveys of attention mechanisms and their applications are
provided by Chaudhari et al. \cite{Chaudhari-etal-2019}, Niu et al.
\cite{Niu-Zhong-Yu-21}, Lin et al. \cite{Lin-Wang-etal-21}. Returning to
explanation models, we can note that the idea to visualize attention weights
jointly with an image instance directly leads to application of attention to
solving the explanation or interpretation tasks. As a result, several studies
discuss pros
\cite{Hickmann-etal-21,Li-Zhang-Chen-21,Patro-etal-20,Rojat-etal-21,Skrlj-etal-20,Wiegreffe-Pinter-19}
and cons \cite{Jain-Wallace-19,Serrano-Smith-19} of the attention application
to problems of explanation. However, most results concerning with explanation
by means of the attention mechanism are devoted to image-based analysis and
NLP when inputs of a black-box model are represented as images or textual
data, respectively. To the best of our knowledge, there are a few explanation
methods (see, for example, \cite{Chang-Caruana-etal-21}) using the attention
mechanism for explaining tabular data.

Another important problem, which can be met in explanation, is a lack of
efficient methods for identifying interactive effects or interactions between
features when only combinations of two and larger features explain predictions
of a black-box model. This problem has been partially solved in NAM and its
modifications \cite{Agarwal-etal-20,ONeill-etal-21,Yang-Zhang-Sudjianto-20}
where pairs of features can be fed to separate subnetworks and provide shape
functions of two variables as outputs of these networks. The main obstacle for
implementing NAM with pairs of features is that a huge number of subnetworks
should be trained. Another way for considering interactions is to use EBM
\cite{Nori-etal-19}. However, shape functions in EBM are gradient-boosted
ensembles of trees such that each tree deals with a single feature or a pair
of features. This also leads to a large number of trees for training.
Moreover, EBM as an extension of the gradient boosting machine uses the greedy
strategy which may lead to non-optimal solutions and incorrect results.

Taking into account the above, we propose a new attention-like explanation
method which overcomes these difficulties and deals with tabular data. It uses
many neural networks similar to NAM as its part and elements of a specific
attention mechanism. That is why we use the term \textquotedblleft
attention-like\textquotedblright\ to characterize the proposed method. The
method is implemented in the form of a system which is called AFEX
(Attention-like Feature EXplanation) and consists of several components,
including, many neural networks (a set of one-feature neural subnetworks) for
feature transformation, a specific scheme of the transformed features
combining with the target values by means of the attention mechanism. The
system parameters are trained by using the end-to-end gradient descend
algorithm. In contrast to the original attention mechanism, we propose to
combine not feature vectors with target values, but the data vectors
corresponding to a single feature. In other words, a matrix of data (each
instance is the corresponding columns) is transposed before applying the
attention mechanism. The idea of this transposition is crucial and aims to
assign weights to features which have a more strong connection with the
corresponding target values. We propose the \textit{Feature attention} instead
of the standard \textit{Instance attention}.

AFEX is trained on the whole set of points including points from the dataset
and generated points around the dataset points. If the total explanation
system shows convergence during training for all data, then points from a
small local area around an explanation point of interest from the area
covering the training data can be processed without additional training of the
system. This peculiarity defines the important property of universality.
Moreover, it should be noted that some effects of features may be not detected
on the local area especially, when there are global effects. In this case, it
is better to detect effects on the whole set of data. The above does not mean
that the method can be used only for the global explanation. It can be
successfully applied to the local explanation. But the local explanation is
obtained through the global one to some extent.

An important property of AFEX is that a basis of shape functions for every
feature is constructed such that the final shape function of a feature is
computed as the weighted sum of shape functions from the basis. Here the
weights are obtained by using the attention mechanism. In other words,
attention weights are used not for indicating the feature impact. They are for
getting shape functions from subsets of basis shape functions.

Another idea behind AFEX or its modification for analyzing the pairwise
interactions of features is based on pairwise multiplications of shape
functions corresponding to different features. As a result, we do not
construct many one-feature neural subnetworks plus a huge number of
two-feature neural subnetworks. The pairwise multiplications of feature
vectors are formed at the attention-like scheme.

In order to avoid the subnetwork overfitting, \textquotedblleft a shortcut
connection\textquotedblright\ with the trainable parameter\ is added to every
subnetwork. Finally, we also consider a modification of AFEX, which is based
on an additional surrogate model implemented as a simple neural network and
trained jointly with other components of the system. This idea aims to cope
with using a complex black-box model for getting predictions and to reduce the
inference time of the whole system.

In summary, the following contributions are made in this paper:

\begin{enumerate}
\item We propose a new explanation method and its implementation AFEX for
identifying important features as well as pairwise interactions between
features, which, in contrast to GAMI-Net, is performed without quadratic
growth in the number of neural subnetworks.

\item An important advantage of AFEX is that the whole system is trained
end-to-end only once on an area of data around some central point such that
the choice of a point for explanation in future does not require to train the
neural subnetworks again, i.e., the feature contributions in the form of shape
functions are obtained without additional training.

\item We propose to compute weighted sums of shape functions for every feature
by using the attention-like mechanism. It is shown by numerical examples that
the best attention is the linear regression which weights are recomputed at
each forward pass.

\item Shortcut connections\ with the trainable parameter\ is used to avoid
overfitting of neural subnetworks due to linearization of the basis shape
functions corresponding to unimportant features.

\item We propose a modification of AFEX by adding a surrogate model
implemented as a simple neural network and trained jointly with other
components of the system. The modification improves the explanation system performance.

\item AFEX is illustrated by means of numerical experiments with synthetic and
real data.
\end{enumerate}

The code of the proposed algorithm can be found in https://github.com/andruekonst/afex.

The paper is organized as follows. Related work is in Section 2. A brief
introduction to the attention mechanism and to NAM is given in Section 3. AFEX
without considering pairwise interactions of features is introduced in Section
4. Pairwise interactions of features are considered in Section 5. The
modification of AFEX with the additional surrogate model is studied in Section
6. Numerical experiments with synthetic data and real data are given in
Section 7. Concluding remarks can be found in Section 8.

\section{Related work}

\textbf{Local and global explanation methods.} The need to explain black-box
models led to development of many interesting explanation methods. One of the
important methods is LIME \cite{Ribeiro-etal-2016}. Several modifications of
LIME have been proposed due to its simplicity and clarity, for example, ALIME
\cite{Shankaranarayana-Runje-2019}, LIME-Aleph \cite{Rabold-etal-2019},
SurvLIME \cite{Kovalev-Utkin-Kasimov-20a}, LIME for tabular data
\cite{Garreau-Luxburg-2020}, GraphLIME \cite{Huang-Yamada-etal-2020}, etc.
LIME and its modifications use the perturbation techniques
\cite{Fong-Vedaldi-2019,Vu-etal-2019} which are based on generating many
points around an explained instance and measuring how predictions are changed
when features are altered \cite{Du-Liu-Hu-2019}. The second important method,
which motivated to develop many modifications, is SHAP
\cite{Strumbel-Kononenko-2010}. It is based on a game-theoretic approach using
Shapley values \cite{Lundberg-Lee-2017}. Several modifications and extensions
of SHAP have been also proposed, for example, FastSHAP\cite{Jethani-etal-21},
Neighbourhood SHAP \cite{Ghalebikesabi-etal-21}, SHAFF \cite{Benard-etal-21},
X-SHAP \cite{Bouneder-etal-20}, ShapNets \cite{Wang-Wang-Inouye-21}, etc.

It should be noted that LIME and SHAP are only two approaches among many
interesting methods which are successfully used in many applications. We do
not consider many methods of counterfactual explanations
\cite{Wachter-etal-2017}, visual explanations \cite{Hendricks-etal-2018} and
other various methods which have been proposed to solve the problem of the
machine learning model prediction explanation because LIME and SHAP can be
regarded as a basis for developing several methods using GAMs for explanation.
Comprehensive reviews providing details and peculiarities of most explanation
methods can be found in survey papers
\cite{Adadi-Berrada-2018,Arrieta-etal-2020,Bodria-etal-21,Carvalho-etal-2019,Guidotti-2019,Li-Xiong-etal-21,Rudin-2019,Rudin-etal-21}%
.

\textbf{GAMs and NAMs in explanation problems}. Due to flexibility and
generality of GAM, this model has been successfully used for developing new
local and global explanation models. In particular, explanation models based
on the gradient boosting machines \cite{Friedman-2001} to produce GAMs were
proposed by \cite{Konstantinov-Utkin-21,Lou-etal-12,Zhang-Tan-Koch-etal-19}.
Gradient boosting machines and GAMs were also used to construct the EBM
\cite{Chang-Tan-etal-2020,Nori-etal-19}. A differentially private version of
EBMs called DP-EBM was proposed by Nori et al. \cite{Nori-etal-21}.

A linear combination of neural networks implementing shape functions in GAM is
a basis for NAMs \cite{Agarwal-etal-20} which sufficiently extend the
available explanation methods and, in fact, open a door for constructing a new
class of methods. Similar approaches using neural networks to construct GAMs
and to perform shape functions are the basis of methods called GAMI-Net
\cite{Yang-Zhang-Sudjianto-20} and AxNNs \cite{Chen-Vaughan-etal-20}. An
architecture called the regression network which can be also regarded as a
modification of NAM is proposed by O'Neill et al. \cite{ONeill-etal-21}.

A method of applying GAM to recommendation system and taking into account
feature interactions was proposed by Guo et al. \cite{Guo-Su-etal-20} and is
called GAMMLI. The neural GAM called NODE-GAM and neural GA$^{2}$M
(NODE-GA$^{2}$M) models were proposed by Chang et al.
\cite{Chang-Caruana-etal-21} as an extension of NAM to deal with large
datasets and to capture abrupt changes which cannot be detected by NAM. The
models are based on using the neural oblivious decision trees
\cite{Popov-etal-19}. A question of using a sparse GAM representation is
studied in \cite{Chang-Tan-etal-2020}. Application of NAM and the sparse GAM
to solving the survival analysis problem under censored training data was also
considered in \cite{Utkin-Satyukov-Konstantinov-2021}.

\section{Preliminary}

\subsection{Basics of the attention mechanism}

It is pointed out in \cite{Chaudhari-etal-2019,Zhang2021dive} that the
original idea of attention can be understood from the statistics point of view
applying the Naradaya-Watson kernel regression model
\cite{Nadaraya-1964,Watson-1964}. Given $n$ training instances
$S=\{(\mathbf{x}_{1},y_{1}),(\mathbf{x}_{2},y_{2}),...,(\mathbf{x}_{n}%
,y_{n})\}$, in which $\mathbf{x}_{i}=(x_{i1},...,x_{id})\in\mathbb{R}^{d}$
represents a feature vector involving $d$ features and $y_{i}\in\mathbb{R}$
represents the regression outputs, the task of regression is to construct a
regressor $f:\mathbb{R}^{d}\rightarrow\mathbb{R}$ which can predict the output
value $y$ of a new observation $\mathbf{x}$, using available training data
$S$. The similar task can be formulated for the classification problem.

The original idea behind the attention mechanism is to replace the simple
average of outputs $y^{\ast}=n^{-1}\sum_{i=1}^{n}y_{i}$ for estimating the
regression output $y$ corresponding to a new input feature vector $\mathbf{x}$
with the weighted average in the form of the Naradaya-Watson regression model
\cite{Nadaraya-1964,Watson-1964}:%
\begin{equation}
y^{\ast}=\sum_{i=1}^{n}\alpha(\mathbf{x},\mathbf{x}_{i})y_{i},
\end{equation}
where weight $\alpha(\mathbf{x},\mathbf{x}_{i})$ conforms with relevance of
the $i$-th training instance to the vector $\mathbf{x}$.

In other words, according to the Naradaya-Watson regression model, to estimate
the output $y$ of an input variable $\mathbf{x}$, training outputs $y_{i}$
given from a dataset weigh in agreement with the corresponding input
$\mathbf{x}_{i}$ locations relative to the input variable $\mathbf{x}$. The
closer an input $\mathbf{x}_{i}$ to the given variable $\mathbf{x}$, the
greater the weight assigned to the output corresponding to $\mathbf{x}_{i}$.

One of the original forms of weights is defined by a kernel $K$ (the
Nadaraya-Watson kernel regression \cite{Nadaraya-1964,Watson-1964}), which can
be regarded as a scoring function estimating how vector $\mathbf{x}_{i}$ is
close to vector $\mathbf{x}$. The weight is written as follows
\cite{Zhang2021dive}:
\begin{equation}
\alpha(\mathbf{x},\mathbf{x}_{i})=\frac{K(\mathbf{x},\mathbf{x}_{i})}%
{\sum_{j=1}^{n}K(\mathbf{x},\mathbf{x}_{j})}.
\end{equation}

The above expression is an example of weights in nonparametric attention
\cite{Zhang2021dive}. In terms of the attention mechanism
\cite{Bahdanau-etal-14}, vector $\mathbf{x}$, vectors $\mathbf{x}_{i}$ and
outputs $y_{i}$ are called as the query, keys and values, respectively. Weight
$\alpha(\mathbf{x},\mathbf{x}_{i})$ is called as the attention weight.
Therefore, the standard attention applications are often represented in terms
of queries, keys and values, and the attention weights are expressed through
these terms.

Generally, weights $\alpha(\mathbf{x},\mathbf{x}_{i})$ can be extended by
incorporating learnable parameters. For example, if we denote $\mathbf{q=W}%
_{q}\mathbf{x}$ and $\mathbf{k}_{i}\mathbf{=W}_{k}\mathbf{x}_{i}$ referred to
as the query and key embeddings, respectively, then the attention weight can
be represented as:%

\begin{equation}
\alpha(\mathbf{x},\mathbf{x}_{i})=\text{softmax}\left(  \mathbf{q}%
^{\mathrm{T}}\mathbf{k}_{i}\right)  =\frac{\exp\left(  \mathbf{q}^{\mathrm{T}%
}\mathbf{k}_{i}\right)  }{\sum_{j=1}^{n}\exp\left(  \mathbf{q}^{\mathrm{T}%
}\mathbf{k}_{j}\right)  }, \label{Expl_At_12}%
\end{equation}
where $\mathbf{W}_{q}$ and $\mathbf{W}_{k}$ are parameters which are learned,
for example, by incorporating an additional feed forward neural network within
the system architecture.

Several attention weights defining the attention mechanisms have been proposed
for different applications. They can be divided into the additive attention
\cite{Bahdanau-etal-14} and multiplicative or dot-product attention
\cite{Luong-etal-2015,Vaswani-etal-17}. As pointed out by
\cite{Niu-Zhong-Yu-21}, the attention mechanisms can be also classified as
general attention, concat attention, and a location-based attention mechanisms
\cite{Luong-etal-2015}. In particular, the general attention uses learnable
parameters for keys and queries as it is illustrated in \ref{Expl_At_12}
(parameters $\mathbf{W}_{q}$ and $\mathbf{W}_{k}$). The concat attention uses
the concatenation of keys and queries. In the location-based attention, the
scoring function depends only on queries and does not depend on keys. A list
of common attention types can be found in \cite{Niu-Zhong-Yu-21}.

\subsection{Neural additive model}

AFEX has some elements which are similar to NAM \cite{Agarwal-etal-20}.
Therefore, this approach is briefly described. NAM\ can be represented as a
neural network consisting of many subnetworks (one-feature subnetwork) such
that a single feature is fed to each subnetwork. The number of subnetworks is
equal to the number of features in data. The $i$-th subnetwork implements a
function $g_{i}(x)$ which can be regarded as a univariate shape function in
GAM. In other words, NAM implements GAM $y(\mathbf{x})=g_{1}(x_{1}%
)+...+g_{m}(x_{m})$ under condition that functions $g_{i}(x)$ are obtained by
training the whole neural network with the loss function minimizing the
expected difference between the whole network output $y(\mathbf{x}_{j})$ and
the target value $y_{j}$ taken from training data. Subnetworks may have
different structures, but they are jointly trained using backpropagation and
can learn arbitrarily complex shape functions.

Details of the whole neural network training and testing are considered in
\cite{Agarwal-etal-20}.

NAM\ is a very efficient and rather universal method, and ideas behind NAM
lead to new modifications and applications of the explanation method. In
particular, a modification called SurvNAM dealing with censored data in the
framework of survival analysis has been proposed in
\cite{Utkin-Satyukov-Konstantinov-2021}.

To implement the proposed attention-like explanation method, we do not use
separately NAM. The idea of NAM to get functions $g_{i}(x_{i})$ of single
features is used to transform initial feature vectors. In AFEX, we use
subnetworks, transforming features $x_{i}$, as elements of the whole system,
i.e., we train the subnetworks jointly with other elements of the explanation system.

\section{Description of AFEX}

The main idea behind AFEX is to train a universal transparent explanation
system that allows us to produce local and global explanations for any points
of data by using information about the input features as well as about values
{}{}of the target variable. Values {}{}of the target variable are used only
for weighing the shape functions that depend on the input features.

\begin{figure}
[ptb]
\begin{center}
\includegraphics[
height=3.3583in,
width=5.1339in
]%
{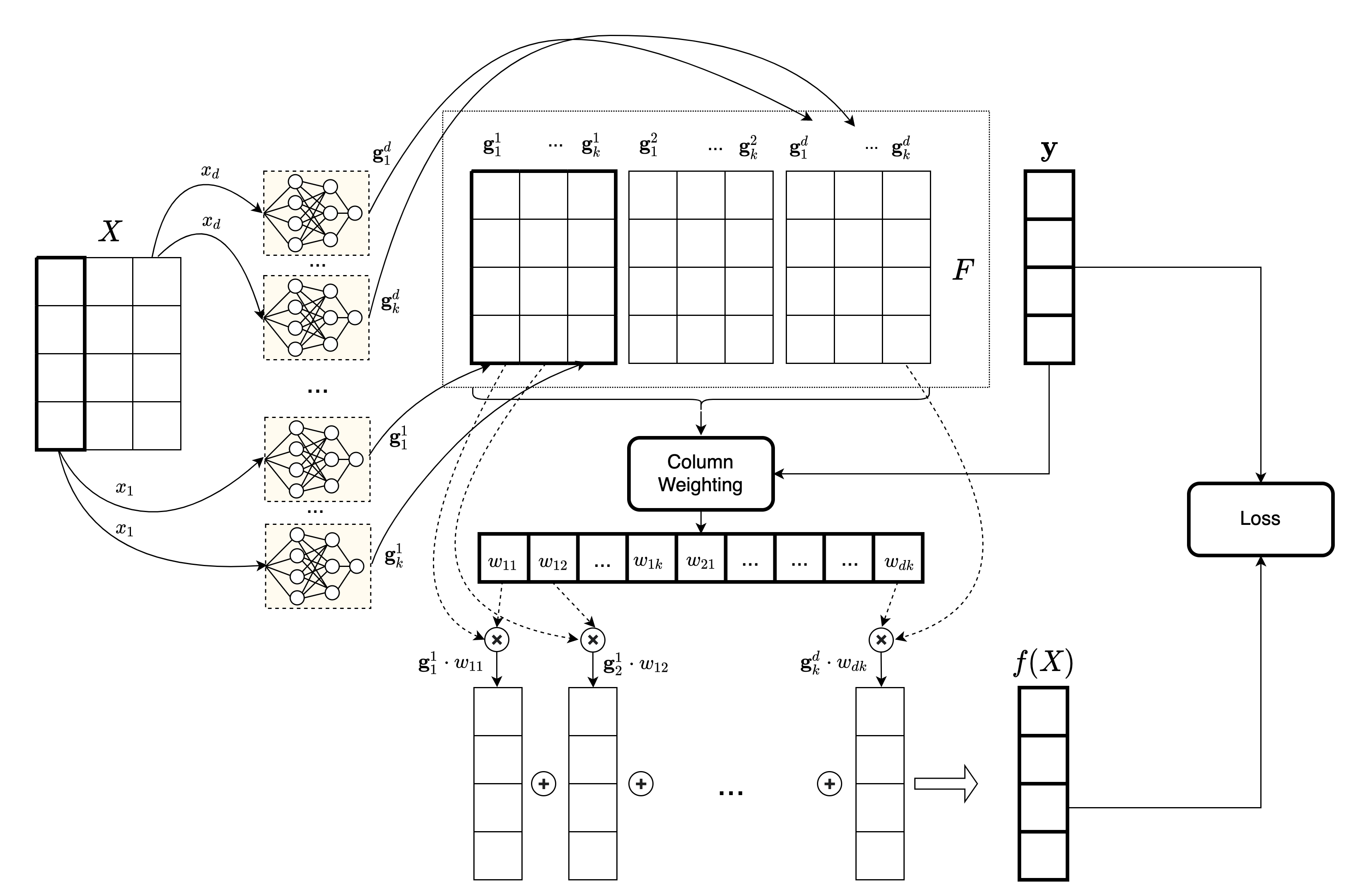}%
\caption{An architecture of the explanation system}%
\label{f:Explan_Syst_1}%
\end{center}
\end{figure}

Let us denote the vector of target values consisting of all outputs defined in
a dataset $S$ as $\mathbf{y}=(y_{1},...,y_{n})$. An architecture of the
proposed explanation system is shown in Fig. \ref{f:Explan_Syst_1}. Here $X$
is the matrix $d\times n$ of input features such that every column of the
matrix contains values of one feature for all instances. Every row contains
features of a single instance. For every feature, we train $k$ different
one-feature neural subnetworks which compute $k$ different shape functions or
shape vectors $\mathbf{g}_{i}^{j}$, $j=1,...,k$, of the same feature $x_{i}$.
Every shape function is a vector of size $n$, i.e., $\mathbf{g}_{i}%
^{j}=\left(  g_{i,1}^{j},...,g_{i,n}^{j}\right)  $, where $g_{i,t}^{j}$ is the
value of the shape function for feature $x_{i}$ of the $t$-th instance
produced by the $j$-th neural subnetwork. It should be noted that the
subnetworks can be identical, but they do not share parameters. By means of
$k$ subnetworks for a single feature, we get $k$ different feature
representations or $k$ different shape functions. The total set of subnetworks
is $k\cdot d$. Here $k$ can be regarded as a tuning parameter. Every $k$ shape
function can also be viewed as a basis of functions such that every function
in the shape function space can be represented as a linear combination of
these basis functions with weights which will be considered below.

The result of the first part of the explanation system is $k\cdot d$ vectors
of data corresponding to $d$ features, or $k\cdot d\cdot n$ values of shape
functions, or matrix $F$ of the size $n\times(k\cdot d)$, where each column
corresponds to the shape function $\mathbf{g}_{i}^{j}$. Matrix $F$ is obtained
by concatenating of $k\cdot d$ columns.

One of the important ideas behind AFEX is to find attention weights of all
shape functions $\mathbf{g}_{i}^{j}$ which establish the relationship between
the shape function values and the target values $\mathbf{y}$, i.e., between
$g_{i,t}^{j}$ and $y_{t}$, $t=1,...,n$. In fact, subsets of the weights can be
regarded as the basis function weights. If the weights were computed, we would
find weighted sums of shape functions which show the feature behavior and
their importance. So, the next step is to compute weights of the matrix $F$ columns.

Weights can be computed by different ways:

\begin{itemize}
\item By means of attention which is based on the dot-product with the softmax
operation $\mathrm{softmax}\left(  \mathbf{y}^{\mathrm{T}}\cdot\mathbf{g}%
_{i}^{j}\right)  $ without a weighted matrix.

\item By means of the correlation coefficient.

\item By means of the linear regression.
\end{itemize}

All these approaches can be viewed as special forms of the attention
mechanism. It is interesting to point out that the attention is applying to
features but not to instances, i.e., we use the \textit{Feature Attention}
instead of the \textit{Instance Attention}.%

Suppose that we have computed the weights of the matrix $F$ columns. The next
step aims to get the prediction $f(\mathbf{x}_{i}\mathbf{)}$ which is an
approximation of $y_{i}$ for all $i=1,...,n$. It is performed by multiplying
columns of matrix $F$ by the obtained weights $w_{ij}$ and then summing up
results of the products.

The error backpropagation is used to train the whole explanation system. The
loss function, for example, mean squared error (MSE), is defined as a distance
between the predicted values (predictions of the explanation model)
$f(\mathbf{x}_{i}\mathbf{)}$ {}{}and the target variable $y_{i}$. A batch is
generated around a randomly selected point at each iteration of training, and
the predicted values {}{}of the target variable are calculated using AFEX.
Since the attention or linear regression weights are calculated in the
differentiable way, the AFEX training is similar to the neural network
training. As a result, if the training process converges, then, for any point
of the training set, the weighted set of one-variable functions will provide a
desirable approximation of the explained black-box model.

The next question is how to implement $k\cdot d$ neural subnetworks with one
input feature. On the one hand, they should be large enough to approximate the
function of the black-box model. On the other hand, they should not lead to
overfitting. It is also important that each basic neural subnetwork should not
have a negative effect on the training process of other subnetworks at initial
stages of training. This implies that the initialization function has to be as
simple as possible, for example, the linear one. The simplest way is to
pre-train each neural subnetwork so that it approximates a linear function,
but further training of the neural network may be difficult in this case.%

\begin{figure}
[ptb]
\begin{center}
\includegraphics[
height=1.3513in,
width=3.0908in
]%
{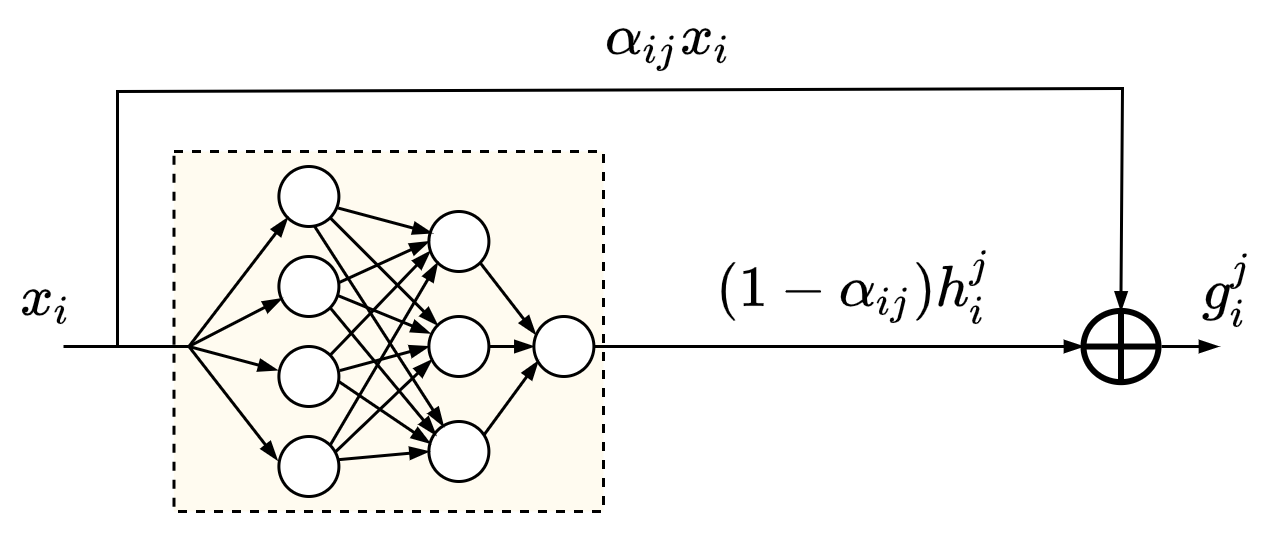}%
\caption{A scheme of the shortcut connection}%
\label{f:shortcut_con_a}%
\end{center}
\end{figure}

Therefore, an alternative idea is proposed. According to the idea, a
\textquotedblleft shortcut connection\textquotedblright\ is added to every
subnetwork as shown in Fig. \ref{f:shortcut_con_a}. In contrast to the
original shortcut connection used in the well-known ResNet
\cite{He-Zhang-Ren-Sun-2016}, we replace the subnetwork with the linear
combination of the linear function $x_{i}$ and the subnetwork output
$h_{i,t}^{j}(x_{i})$ with the combination parameter $\alpha_{ij}$ such that
the parameter $\alpha_{ij}$ is trainable. Its initial value is proposed to
take as $0.9$ to minimize the neural subnetwork impact, which is $0.1$, and to
maximize the linear function impact. This structure with the shortcut
connection provides the shape function $g_{i,t}^{j}=(1-\alpha_{ij})h_{i,t}%
^{j}+\alpha_{ij}x_{t,i}$ for every instance with index $t=1,...,n$. During
training, the combination parameter $\alpha_{ij}$ may be changed, and the
model may implement an arbitrary complex function. If there exists some prior
knowledge about the function implemented by each subnetwork, then a
regularization can be used to realize functions which are close to the linear function.

It is important to note that parameters of AFEX in Fig. \ref{f:Explan_Syst_1},
including parameters of $k\cdot d$ neural subnetworks, weights of shape
functions, linear combination parameters $\alpha_{ij}$, are jointly trained
for the whole system by using the end-to-end learning process.

\subsection{Attention-like differentiable regression}

Let us consider an important part of AFEX which aims to compute weights of the
matrix $F$ columns.

We propose to use a specific attention-like differentiable regression for
computing the weights. Let us discuss other attention mechanisms which could
be used in the proposed model, but are not used due to the following reasons.
We write the scoring functions \textquotedblleft score\textquotedblright%
\ estimating how vector $\mathbf{y}$ is close to vector $\mathbf{g}_{i}^{j}$
and used in the attention operations in terms of our task.

\begin{itemize}
\item The general attention \cite{Luong-etal-2015}: $\mathrm{score}%
(\mathbf{y},\mathbf{g}_{i}^{j})=\mathbf{y}^{\mathrm{T}}W_{a}\mathbf{g}_{i}%
^{j}$. If matrix $W_{a}$ of the attention parameters is trainable, then a
change of an order of instances in a batch leads to change of weights which
become incorrect.

\item The additive attention \cite{Bahdanau-etal-14}: $\mathrm{score}%
(\mathbf{y},\mathbf{g}_{i}^{j})=\tanh\left(  W_{a}[\mathbf{y},\mathbf{g}%
_{i}^{j}]\right)  $. Similarly to the general attention, results may be
incorrect if an order of points changes.

\item The dot-product attention \cite{Luong-etal-2015,Vaswani-etal-17}:
$\mathrm{score}(\mathbf{y},\mathbf{g}_{i}^{j})=\mathbf{y}^{\mathrm{T}%
}\mathbf{g}_{i}^{j}$. Larger weights will be assigned to those features
(columns of $F$) that have larger norms.

\item The content-based attention \cite{Graves-etal-14}: $\mathrm{score}%
(\mathbf{y},\mathbf{g}_{i}^{j})=\frac{\mathbf{y}^{\mathrm{T}}\mathbf{g}%
_{i}^{j}}{\left\Vert \mathbf{y}\right\Vert \left\Vert \mathbf{g}_{i}%
^{j}\right\Vert }$. Weights depend on bias $b$, that is, adding a bias to the
vector $\tilde{y}_{j}=y_{j}+b$ will change the weights. In AFEX, the training
process is carried out on different local parts of the data domain, in which
biases of the target variable are different.

\item A procedure similar to the content-based attention with preliminary
transformation of input vectors, where the average vector $\mathbf{\bar{y}%
}^{\mathrm{T}}$ (or $\mathbf{\bar{g}}_{i}^{j}$) calculated over elements of
each vector is subtracted from each vector $\mathbf{y}^{\mathrm{T}}$ (or
$\mathbf{g}_{i}^{j}$):
\begin{equation}
\mathrm{score}(\mathbf{y},\mathbf{g}_{i}^{j})=\frac{\left(  \mathbf{y}%
^{\mathrm{T}}-\mathbf{\bar{y}}^{\mathrm{T}}\right)  \left(  \mathbf{g}_{i}%
^{j}-\mathbf{\bar{g}}_{i}^{j}\right)  }{\left\Vert \mathbf{y}^{\mathrm{T}%
}-\mathbf{\bar{y}}^{\mathrm{T}}\right\Vert \left\Vert \mathbf{g}_{i}%
^{j}-\mathbf{\bar{g}}_{i}^{j}\right\Vert } \label{Pearson_correl}%
\end{equation}

This is the Pearson linear correlation coefficient. If there are correlated
features, then their improtance in significantly increases while the
importance of other features may be significantly decreased.
\end{itemize}

In order to implement the attention-like regression for computing weights, we
solve the following optimization problem
\begin{equation}
\mathbf{w}_{opt}=\left(  \arg\min_{\mathbf{w}}\left\Vert F\mathbf{w}%
-\mathbf{y}\right\Vert ^{2}\right)  ^{\mathrm{T}}, \label{Expl_At_44}%
\end{equation}
where matrix $F$ is non-square (the number of rows $n$ larger than the number
of columns $k\cdot d$, i.e., $n>k\cdot d$); $\mathbf{w}\in\mathbb{R}^{k\cdot
d}$ is the vector of weights of the matrix $F$ columns; $\mathbf{w}_{opt}$ is
the vector of optimal weights by given $F$ and $\mathbf{y}$.

One of the solution algorithms is the following:

\begin{enumerate}
\item Rank $r$ of matrix $F$ is determined.

\item If the rank is equal to the number of columns, then the QR decomposition
is computed, for example, by using the PyTorch library which supports
gradients computation through the block.

\item Otherwise, the problem is replaced with the problem
\begin{equation}
\min_{w}\left\Vert \left(  F^{\mathrm{T}}F+\lambda I\right)  \mathbf{w}%
-F^{\mathrm{T}}\mathbf{y}\right\Vert ^{2},
\end{equation}
which is solved by means of the QR decomposition (we use in numerical
experiments $\lambda=0.1$).
\end{enumerate}

Another question is how to compute weights in \textquotedblleft Column
weighting\textquotedblright\ (see Fig. \ref{f:Explan_Syst_1}) in case of the
linear regression. To take into account the bias which differs for each new
explanation area, matrix $\tilde{F}=(F;I)$ is constructed by adding a unit
column to matrix $F$. Further, optimal weights $\mathbf{w}_{opt}=\arg
\min_{\mathbf{w}}\left\Vert F\mathbf{w}-\mathbf{y}\right\Vert ^{2}$ are
assigned to columns corresponding to features $F_{(i)}=\mathbf{g}_{i}^{j}$
with the last weight corresponding to the bias.

If we replace the calculation of weights in the attention mechanism using the
softmax score with the calculation of linear regression coefficients, then it
turns out that finding the weights $\mathbf{w}$ by the above method followed
by weighting the input vectors represents the attention-like linear
regression. If the weight vector $\mathbf{w}$ in the original attention
mechanism as the Naradaya-Watson regression model
\cite{Nadaraya-1964,Watson-1964} ($\mathbf{y}=\sum_{i}\mathbf{w}_{i}%
\mathbf{g}_{i}^{j}$) is defined as $\mathbf{w}=$softmax$(\mathrm{score}%
(\mathbf{y},\mathbf{g}_{i}^{j}))$, then the attention-like linear regression
is defined in a similar way, except the weights are determined by solving the
problem (\ref{Expl_At_44}).

\begin{figure}
[ptb]
\begin{center}
\includegraphics[
height=2.6151in,
width=3.8925in
]%
{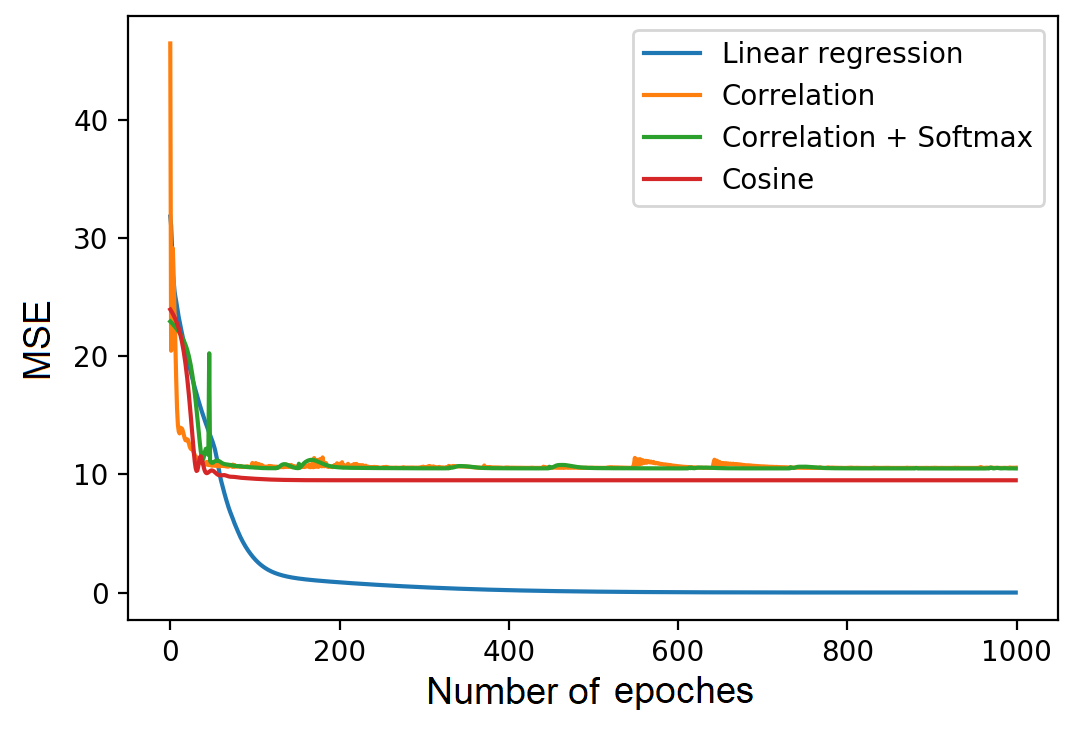}%
\caption{Comparison of different approaches to the column weighting}%
\label{f:weighting_comparison}%
\end{center}
\end{figure}

To illustrate advantages of the attention-like linear regression, we perform
the following experiment. Typical dependencies of the MSE as a function of the
number of epochs for different methods of the weight computation are depicted
in Fig. \ref{f:weighting_comparison}. We compare the following variants
depicted in Fig. \ref{f:weighting_comparison} when the black-box model is the
function $x_{1}^{2}+0.5x_{2}$:

\begin{itemize}
\item Linear regression: the proposed approach, see (\ref{Expl_At_44}).

\item Correlation: see (\ref{Pearson_correl}).

\item Correlation+Softmax: see (\ref{Pearson_correl}) and (\ref{Expl_At_12}).

\item Cosine: see the content-based attention \cite{Graves-etal-14}.
\end{itemize}

It can be seen from Fig. \ref{f:weighting_comparison} that the best
convergence is observed for the linear regression. Similar relationships
between different methods take place for other black-box model functions.%

\begin{figure}
[ptb]
\begin{center}
\includegraphics[
height=1.217in,
width=4.5483in
]%
{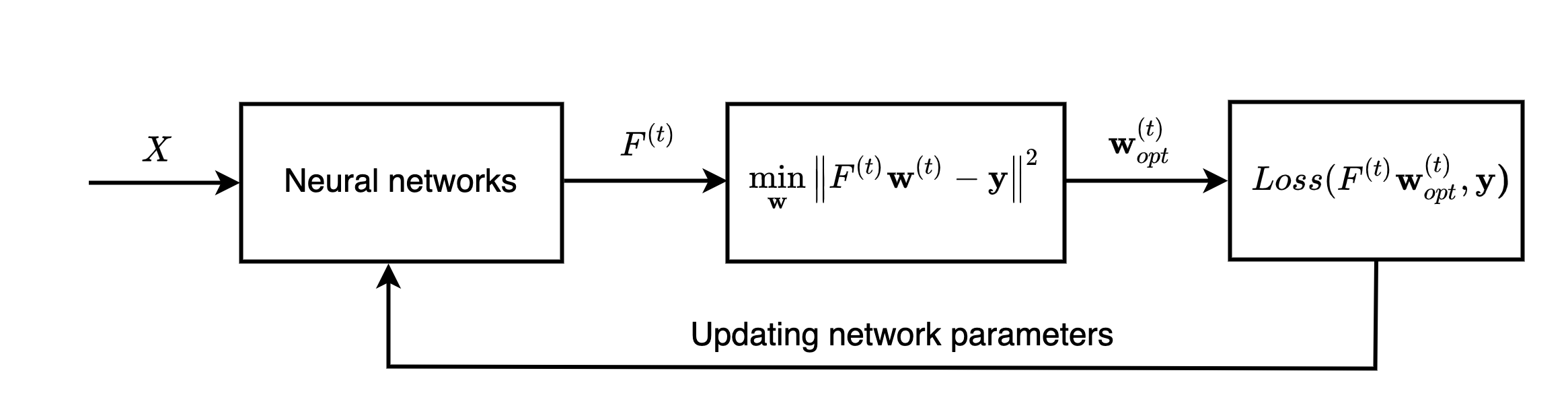}%
\caption{An enlarged scheme of the system training process}%
\label{f:enlarged_scheme}%
\end{center}
\end{figure}

We have to point out that the linear regression is solved for every batch of
data during training the whole system. In other words, trainable parameters of
the system (weights of neural subnetworks) are modified after propagation a
single batch of data, and new values of vectors $\mathbf{g}_{1}^{1,(t)}%
,...,\mathbf{g}_{k}^{d,(t)}$ are computed, and they form matrix $F^{(t)}$,
where $t$ denotes the training iteration number. New weights $\mathbf{w}%
^{(t)}$ are calculated by solving the optimization problem (\ref{Expl_At_44})
under condition that the matrix $F^{(t)}$ is computed at this iteration of the
system training. As a results, $M$ iterations of the system training require
to solve the problem (\ref{Expl_At_44}) $M$ times. The training process is
schematically depicted in Fig. \ref{f:enlarged_scheme} where the $t$-th
iteration of the training is shown. One of the possible loss functions is
\begin{equation}
Loss\left(  F,\mathbf{w},\mathbf{y}\right)  =\left\Vert F\mathbf{w}%
-\mathbf{y}\right\Vert ^{2}.
\end{equation}
%

\subsection{Testing (local explanation)}

\begin{figure}
[ptb]
\begin{center}
\includegraphics[
height=2.8701in,
width=4.7609in
]%
{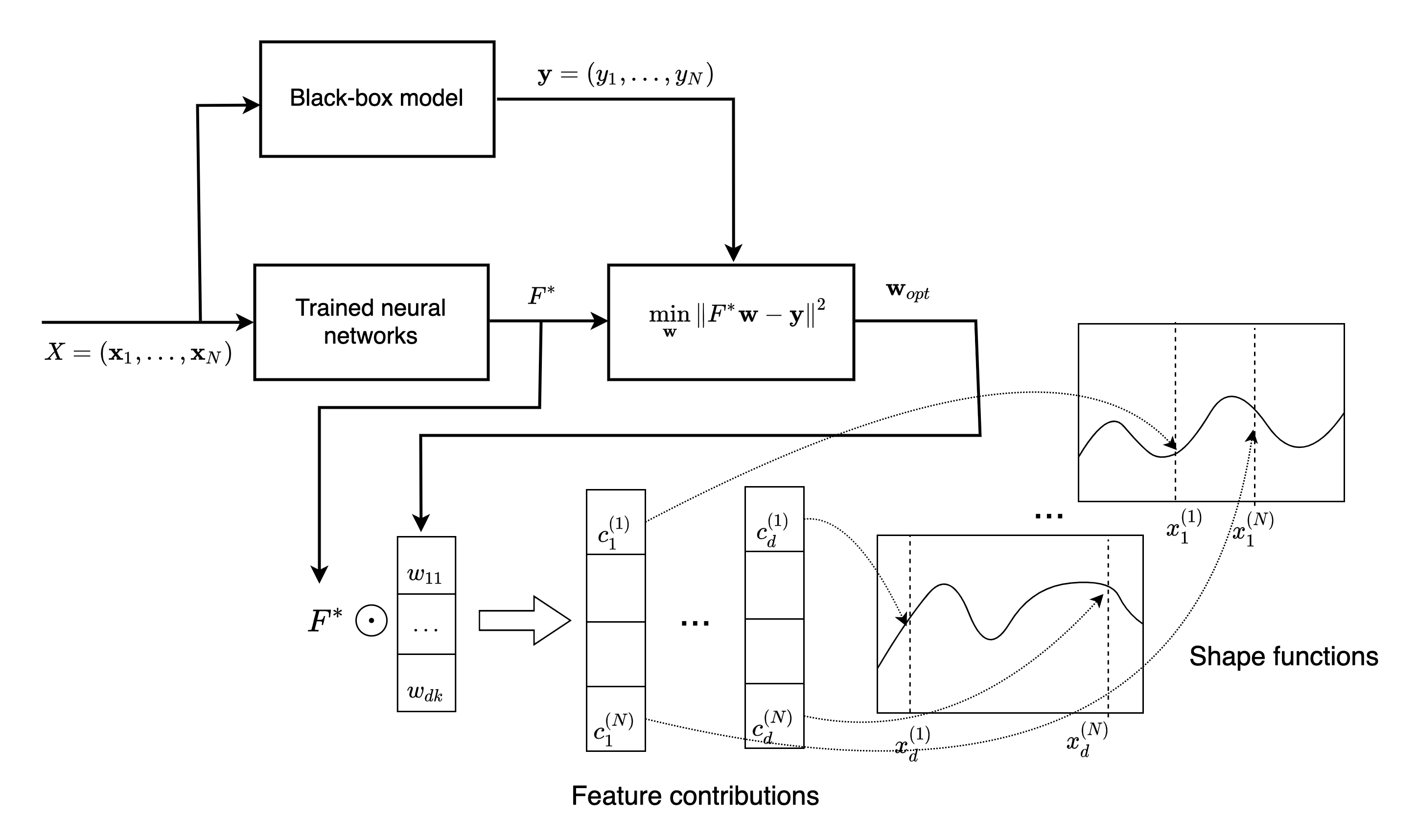}%
\caption{An enlarged scheme of the system testing process (the shape function
construction)}%
\label{f:enlarged_scheme_test}%
\end{center}
\end{figure}

The testing process is schematically depicted in Fig.
\ref{f:enlarged_scheme_test}. $N$ points $\mathbf{x}_{1},...,\mathbf{x}_{N}$
are generated around an explained point. Predictions $y_{1},...,y_{N}$ are
produced by the black-box model for all generated points. Matrix $F^{\ast}$
consisting of $N$ rows and $k\cdot d$ columns is generated by using the
trained neural subnetworks. Matrix $F^{\ast}$ and vector $\mathbf{y}%
=(y_{1},...,y_{N})$ of predictions are used to solve the optimization problem
(\ref{Expl_At_44}) and to get vector $\mathbf{w}_{opt}$ of weights of size
$k\cdot d$. By multiplying every column of matrix $F^{\ast}$ by the
corresponding weight from $\mathbf{w}_{opt}$, we obtain $d$ vectors of feature
contributions of size $N$ which, in turn, produce the shape functions that
show whether the features are important or not.%

\section{An additional surrogate model}

\begin{figure}
[ptb]
\begin{center}
\includegraphics[
height=2.3962in,
width=4.5096in
]%
{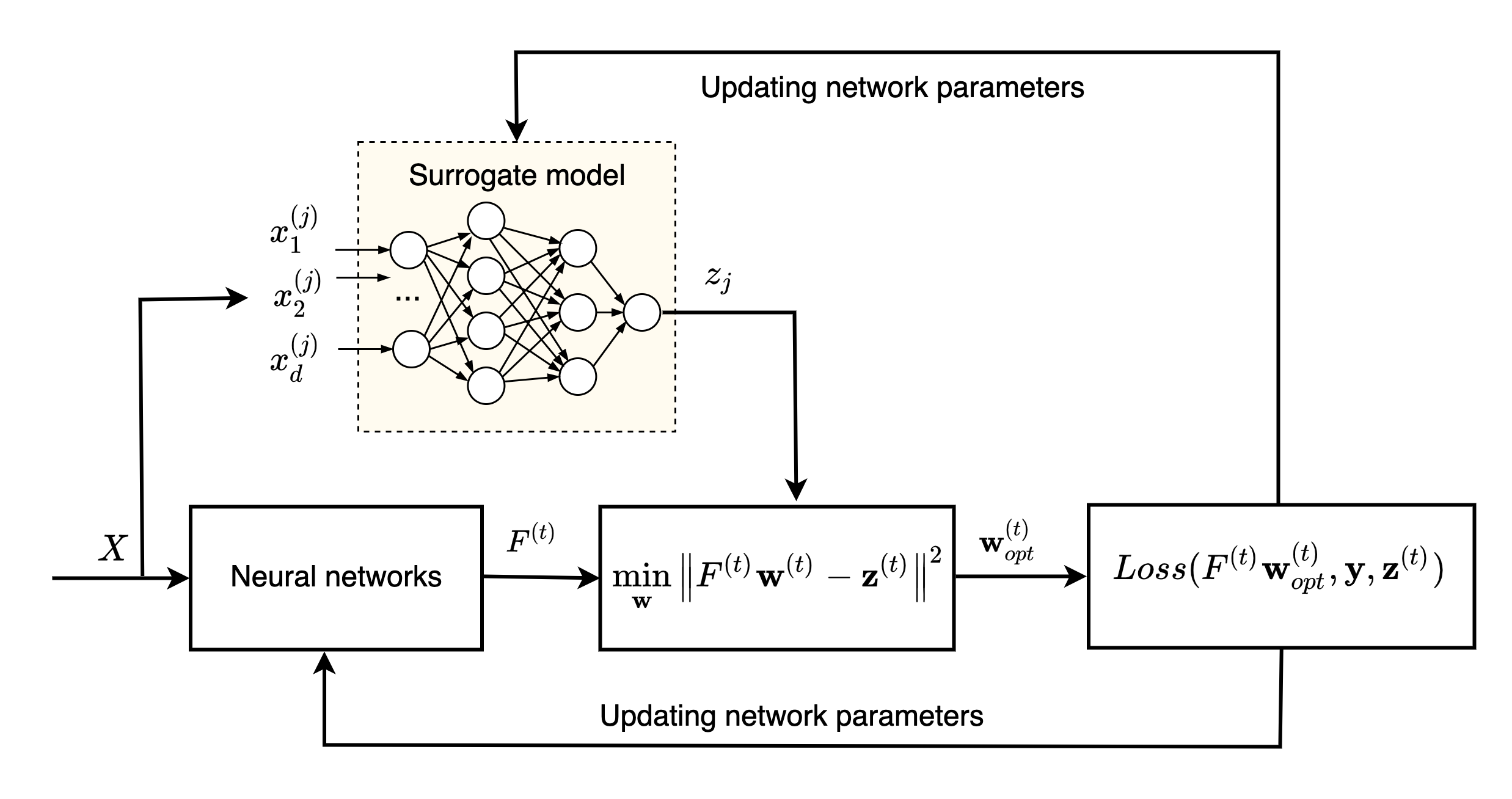}%
\caption{An enlarged scheme of the system training process with the surrogate
model}%
\label{f:surrogate_scheme}%
\end{center}
\end{figure}

In order to improve AFEX, we propose to add a surrogate model which
approximates the black-box model and is trained jointly with the main
explanation model. The surrogate model is implemented as a neural network with
some training parameters (weights of the network connections). An enlarged
scheme of AFEX with the surrogate model is depicted in Fig.
\ref{f:surrogate_scheme}. The surrogate model is trained on the same dataset
as the main explanation model. Its output is vector $\mathbf{z}$ which can be
regarded as an approximation of vector $\mathbf{y}$. In contrast to the
original explanation model, the linear regression block uses vector
$\mathbf{z}$ instead of $\mathbf{y}$. As a result, the joint loss function
takes into account both vectors $\mathbf{z}$ and $\mathbf{y}$. One of the
possible loss functions is
\begin{equation}
Loss\left(  F,\mathbf{w},\mathbf{y},\mathbf{z}\right)  =\left\Vert
F\mathbf{w}-\mathbf{y}\right\Vert ^{2}+\lambda\left\Vert \mathbf{y}%
-\mathbf{z}\right\Vert ^{2},
\end{equation}
where $\lambda$ is the parameter which controls the strength of the second term.

It can be seen from the above loss function that neural networks producing
matrix $F$ and the surrogate model are jointly trained and compensate each
other. Another advantage of using the surrogate model is that we do not need
to use the complex black-box model for getting predictions for all generated
points around an explained point, the surrogate model copes with this task in
a shorter time. Sometimes, there is a dataset, but the black-box model may be
not available. In this case, we train the surrogate model and can investigate
various points of the dataset.%

\section{Identifying pairwise interactions}

One of the main reasons why AFEX is developed is to identify pairwise
interactions among features. It turns out that the model can be simply
modified for identifying the interactions. Interactions are said to exist when
a change in the level of one factor has different effects on the response
variable, depending on the value of the other factor \cite{Dodge-08}. In other
words, the value of one feature modifies the effect of another feature and
vice versa \cite{Boulesteix-etal-14}.

\begin{figure}
[ptb]
\begin{center}
\includegraphics[
height=2.7602in,
width=4.5141in
]%
{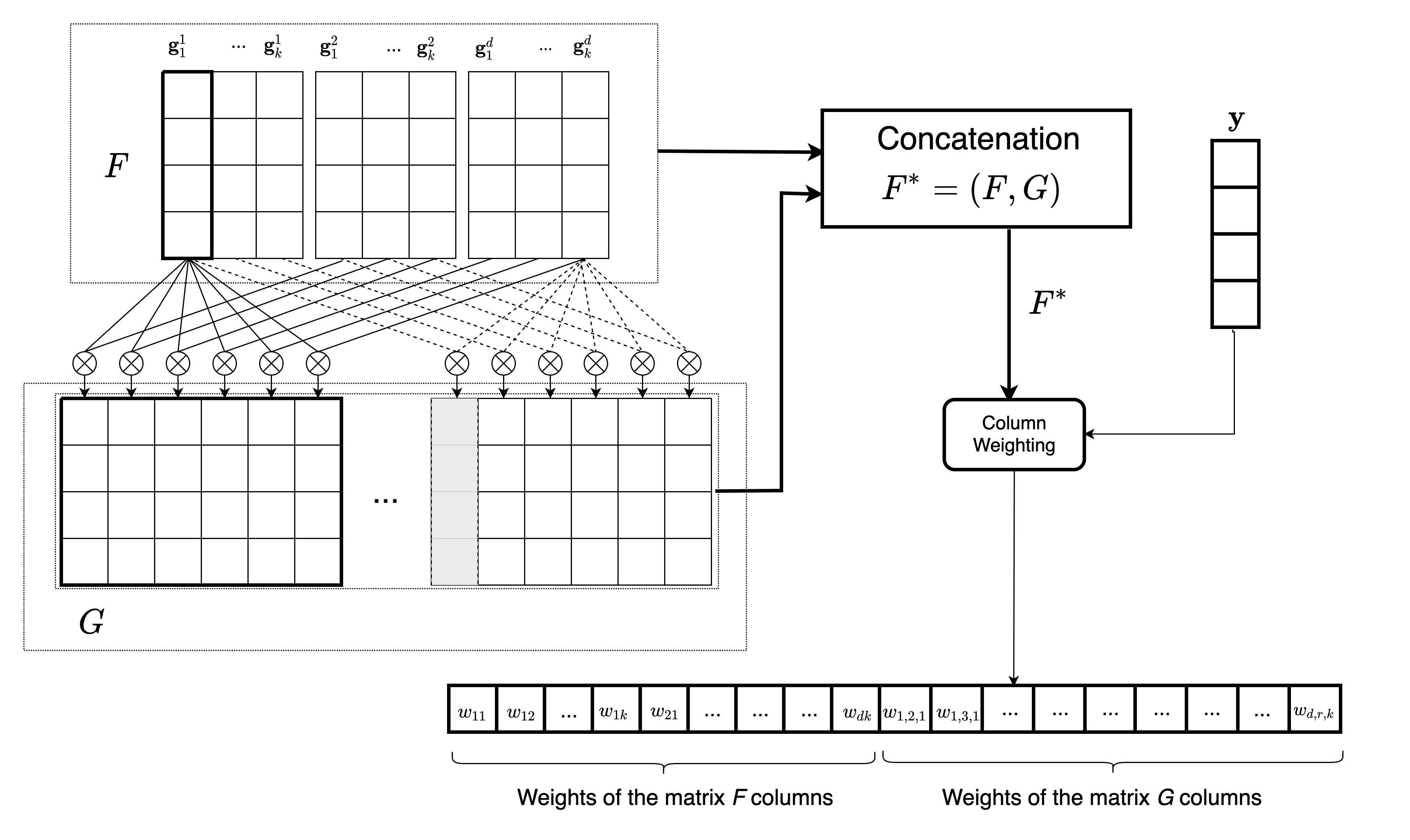}%
\caption{An additional block to the total explanation scheme for identifying
pairwise interactions of features}%
\label{f:scheme_pairwise}%
\end{center}
\end{figure}

One of the most general approaches to identifying pairwise interactions
between two features is the regression approach. According to this approach
the interacted features are represented by multiplying the corresponding
variables \cite{Boulesteix-etal-14}. The idea of the feature multiplication
can be realized by using the proposed model. In order to implement it, we
propose to add a block which computes pairwise products of columns of matrix
$F$. At that, only columns corresponding to different features are multiplied
each other. Columns corresponding to the same features are not multiplied. The
total number of obtained pairs is $k^{2}d(d-1)/2$, i.e., the matrix composed
from the corresponding columns and denoted as $G$ consists of $k^{2}d(d-1)/2$
columns. In sum, the obtained matrix $G$ is concatenated with matrix $F$, and
we get matrix $F^{\ast}=(F,G)$ consisting of $k^{2}d(d-1)/2+kd$ columns and
$n$ rows. This is depicted in Fig. \ref{f:scheme_pairwise}. In fact, Fig.
\ref{f:scheme_pairwise} illustrates a part of the total scheme shown in Fig.
\ref{f:Explan_Syst_1} such that matrix $F$ in Fig. \ref{f:Explan_Syst_1} is
replaced with matrix $F^{\ast}$ and the corresponding weights of columns are
extended. Other parts of the total scheme remain without changes. The model is
trained in the same way, however, the interpretation of results differs.

Each shape function depending on two features $x_{i}$ and $x_{s}$ is computed
as a sum of shape functions of one feature with the corresponding weights
$w_{ij}\cdot\mathbf{g}_{i}^{j}$ plus the sum of shape functions of the second
feature $w_{sl}\cdot\mathbf{g}_{s}^{l}$ and the sum of products of shape
functions with the corresponding weights $w_{i,s,t}\cdot\mathbf{g}_{i}%
^{j}\cdot\mathbf{g}_{s}^{l}$, i.e., $w_{ij}\cdot\mathbf{g}_{i}^{j}+w_{sl}%
\cdot\mathbf{g}_{s}^{l}+w_{i,s,t}\cdot\mathbf{g}_{i}^{j}\cdot\mathbf{g}%
_{s}^{l}$, where $t$ is the number of the pair $(j,l)$. It is important to
note that the number of neural networks to take into account pairwise
interactions does not increase in the proposed approach. This is a very
important difference of the proposed approach from NAM \cite{Agarwal-etal-20}
and its modifications, for example, regression networks \cite{ONeill-etal-21}.
Another advantage of the above approach in comparison with other approaches is
that we get shape functions of pairs of features instead of some coefficients.
A problem, which can be met in standard regression models taking into account
the pairwise interaction, is how to interpret the coefficients of the
corresponding products of variables. It is easy when features are binary.
However, the interpretation in case of real-valued features can be incorrect.
In contrast to these regression models, we obtain shape functions which can be
interpreted in accordance with their changes.%

\section{Numerical experiments}

\subsection{Synthetic data}

\subsubsection{Combination of functions}

To investigate AFEX and its peculiarities, in particular its property that the
training can be carried out once, and explanations can be obtained for
different local areas, the following numerical experiment is performed with
synthetic data. Suppose that the black-box model is a function of two
variables (features) of the form:%
\begin{equation}
I[x_{1}\geq0]\left(  x_{0}\right)  ^{2}+I[x_{1}<0]x_{0}.
\end{equation}

It is equal to the square of the first variable, $x_{0}$, if the second
variable, $x_{1}$, is positive, otherwise, it is equal to the linear function
of the first variable $x_{0}$. Here $I[A]$ is the indicator function taking
value $1$ if $A$ is true. It is important to point out that this function
cannot be approximated by GAM and NAM because these models do not deal with
pairwise interactions.

\begin{figure}
[ptb]
\begin{center}
\includegraphics[
height=2.4584in,
width=2.517in
]%
{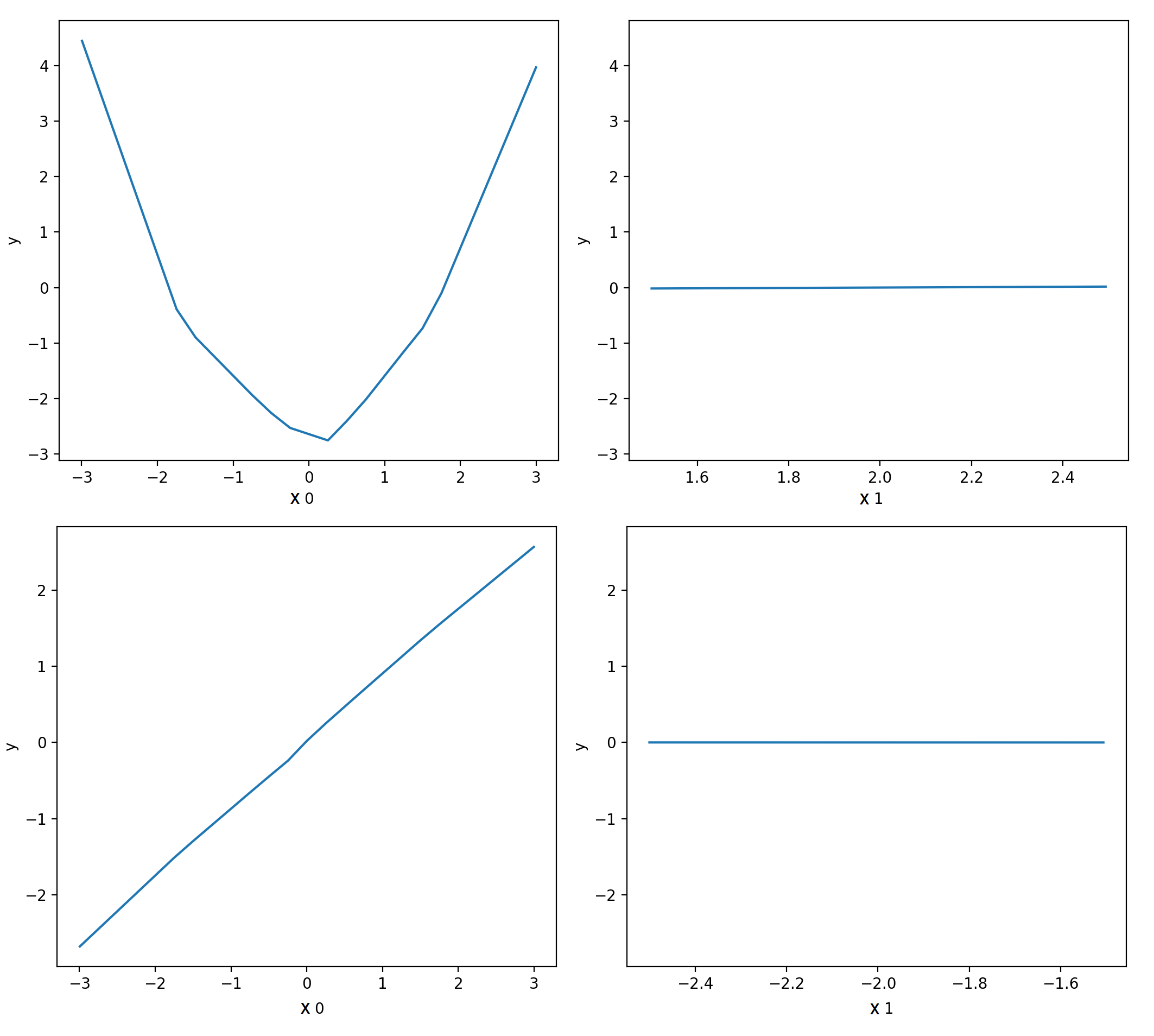}%
\caption{Shape functions explaining the model at two different local areas:
the first and the second rows of pictures correspond to the areas around
$(0,2)$ and $(0,-2)$, respectively}%
\label{f:shape_functions_1}%
\end{center}
\end{figure}

The training process is performed with the batch size equal to $1000$,
instances are randomly generated from the uniform distribution with the
expectation also generated randomly from the normal distribution with the
standard deviation $1$ at each iteration. After training, the obtained model
is used to explain two different local areas around point $(0,2)$ and point
$(0,-2)$. These central points are taken to test two conditions $x_{1}\geq0$
and $x_{1}<0$. Points in the local areas are generated from the uniform
distribution. Resulting shape functions for the first and the second areas
(the first and the second rows of pictures) are shown in Fig.
\ref{f:shape_functions_1}. It can be seen from the first row of pictures in
Fig. \ref{f:shape_functions_1} that the shape function for the first feature
in the first case is close to a parabola, while in the second case, the shape
function represents a linear function. The above implies that the single
explanation model explains various areas of data.%

\begin{figure}
[ptb]
\begin{center}
\includegraphics[
height=2.4025in,
width=2.5647in
]%
{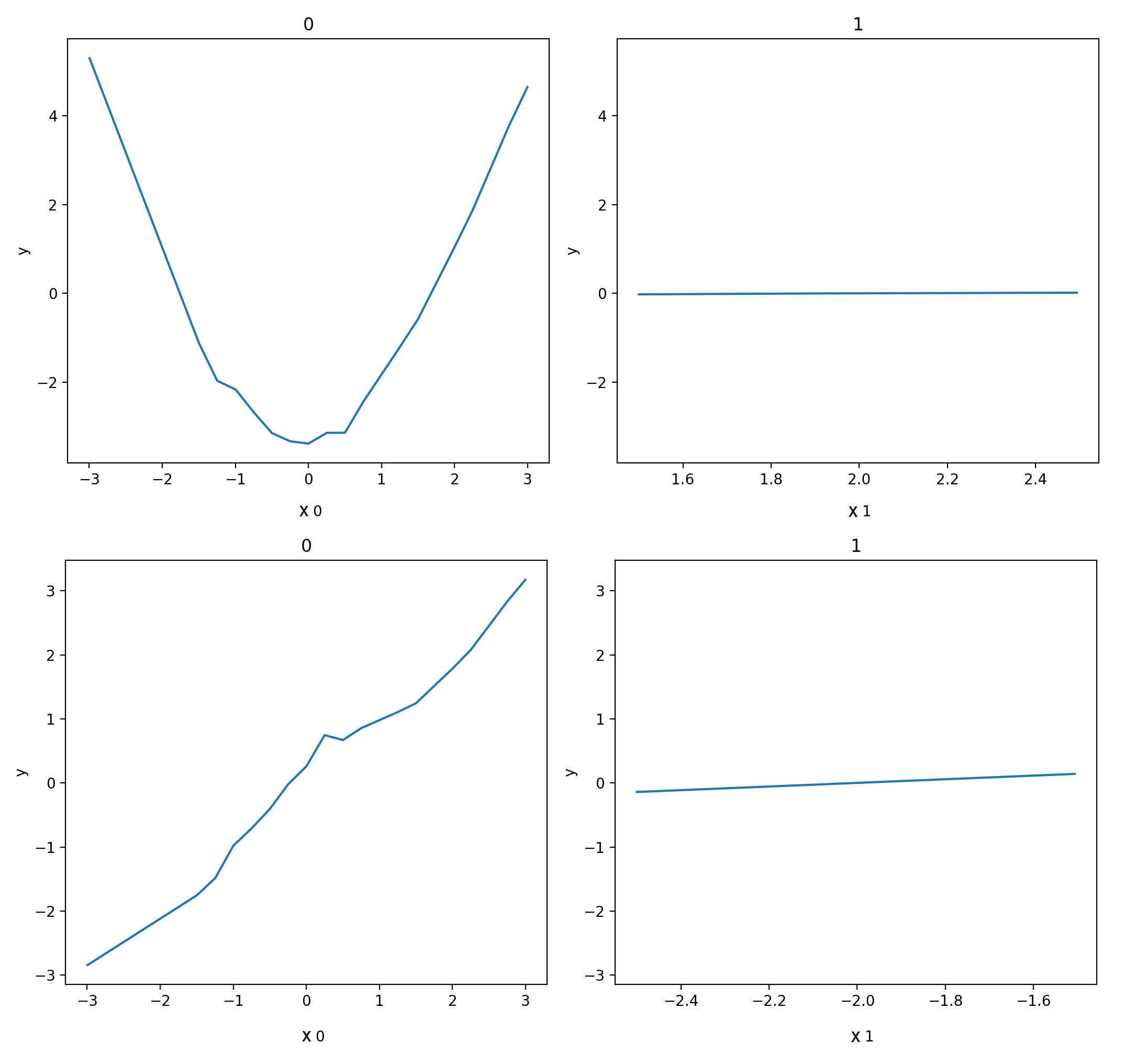}%
\caption{Shape functions explaining the model at two different local areas
obtained by using the modification with surrogate model: the first and the
second rows of pictures correspond to the areas around $(0,2)$ and $(0,-2)$,
respectively}%
\label{f:surrogate_shape_1}%
\end{center}
\end{figure}

To study the modification of the explanation method based on using the
surrogate model, we realized the same experiment with this modification. The
surrogate model is implemented as a neural network consisting of 5 layers such
that each layer has 10 units. The activation function is ReLU. The model has
not been pre-trained. We again explain two different local areas around point
$(0,2)$ and point $(0,-2)$. Results are illustrated in Fig.
\ref{f:surrogate_shape_1}. It can be seen from Fig. \ref{f:surrogate_shape_1}
that the proposed modification of AFEX with the surrogate model provides a
more sensitive results.%

\subsubsection{Chessboard}

The next numerical example demonstrates how AFEX copes with pairwise
interactions of features. The black-box model is a function of five variables,
which significantly depends on two (the first and the second) variables, and
remaining variables do not affect the function values. We use the unimportant
remaining variables because they may complicate the study of pairwise
interactions due to possible overfitting the model. The function of the
explained model is a \textquotedblleft chessboard\textquotedblright\ with
cells of size $2\times2$. Its formal writing is%

\begin{equation}
I\left[  (\sin\left(  x_{0}\cdot\frac{\pi}{2}\right)  >0)\neq(\sin\left(
x_{1}\cdot\frac{\pi}{2}\right)  >0)\right]  .
\end{equation}

\begin{figure}
[ptb]
\begin{center}
\includegraphics[
height=1.8278in,
width=3.134in
]%
{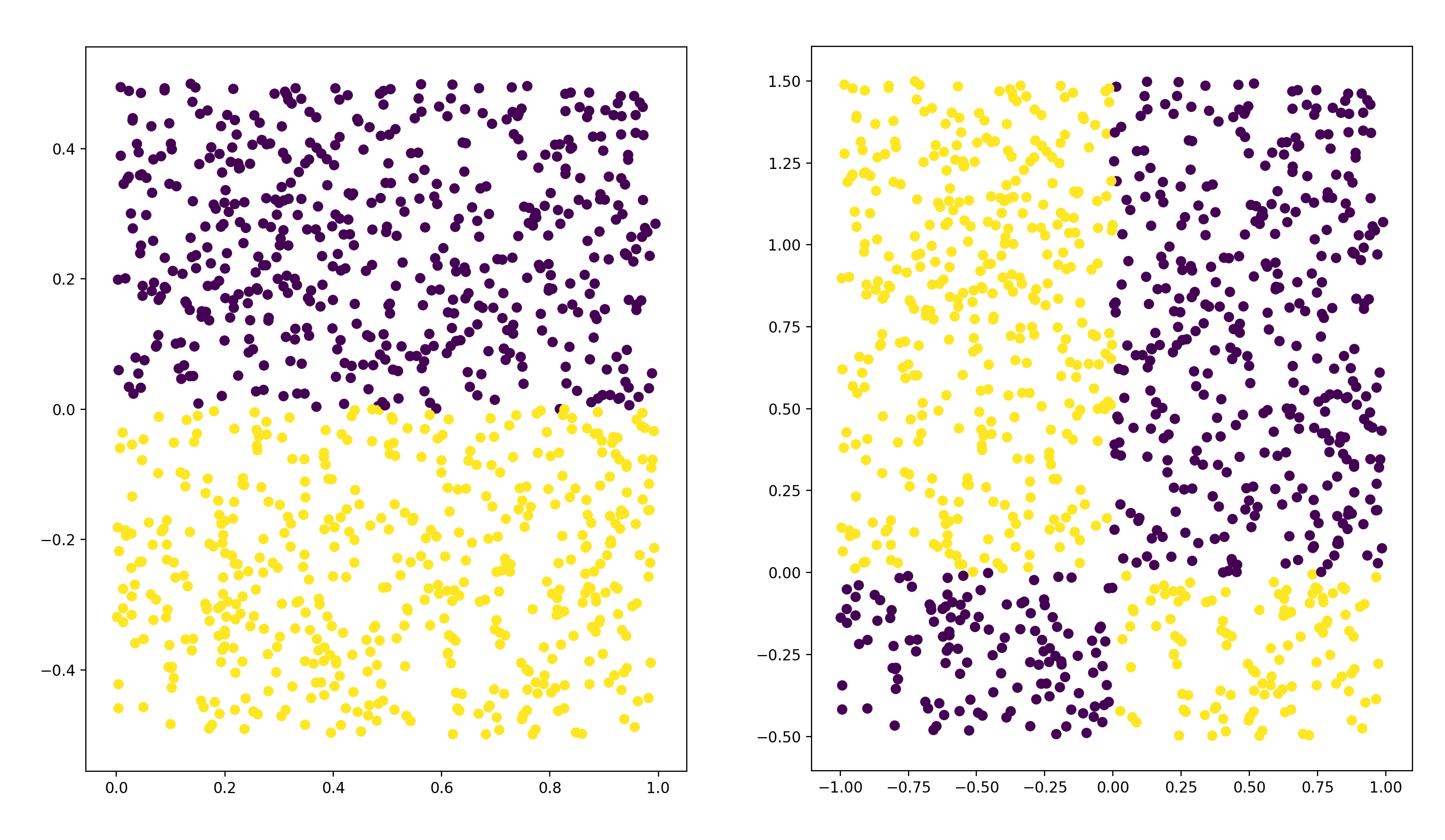}%
\caption{Areas around two points which are close to the horizontal border
between cells (the left picture) and to the diagonal junction of cells (the
right picture)}%
\label{f:target_chess}%
\end{center}
\end{figure}

AFEX is trained by using batches of size $1000$. Instances are randomly
generated from a uniform distribution with a randomly selected center as in
the previous numerical example, but the size of each local area is $0.5$.
Using AFEX, two different points are explained: the first one is close to the
horizontal border between cells (the feature vector $(\frac{1}{2},0,0,0,0)$);
and the second one is close to the diagonal junction of cells (the feature
vector $(0,\frac{1}{2},0,0,0)$) The local area of generated instances is of
size $1$. Target values corresponding these two cases are shown in Fig.
\ref{f:target_chess}. Only two features are used for depicting target values
because it is assumed that other features do not impact on target values.%

\begin{figure}
[ptb]
\begin{center}
\includegraphics[
height=2.5665in,
width=4.5979in
]%
{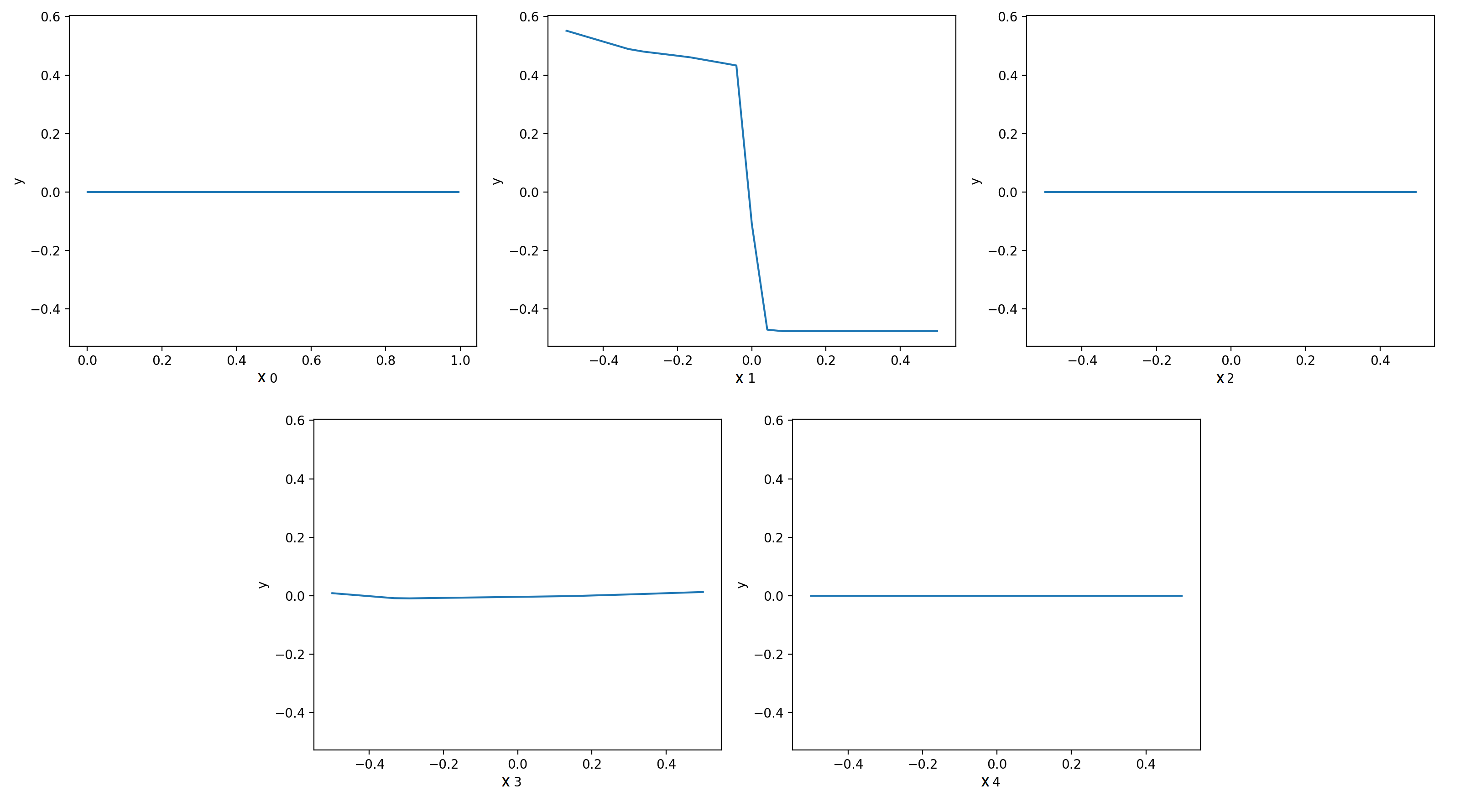}%
\caption{Shape functions of five features for the first local area in the
checkerboard example}%
\label{f:shape_functions_chess_1}%
\end{center}
\end{figure}

Obtained shape functions of five features for the first local area around
point $(\frac{1}{2},0,0,0,0)$ are depicted in Fig.
\ref{f:shape_functions_chess_1}. The strongly marked step-wise shape function
corresponding to the second feature, $x_{1}$, can be seen in Fig.
\ref{f:shape_functions_chess_1}. Indeed, if we look at the left picture in
Fig. \ref{f:target_chess}, then it is obvious that only change of the second
feature leads to change of the chessboard cell. Moreover, this cell change is
carried out at a point $x_{1}=0$ which corresponds to the shape function jump.
It follows from Fig. \ref{f:shape_functions_chess_1} that shape functions of
other features do not impact on the target values. This peculiarity also
corresponds to Fig. \ref{f:target_chess} where it is clearly seen that changes
of feature $x_{0}$ around $x_{0}=0.5$ does not change the cell. Other features
do not impact on the target due to the assumption.%

\begin{figure}
[ptb]
\begin{center}
\includegraphics[
height=3.3908in,
width=4.2069in
]%
{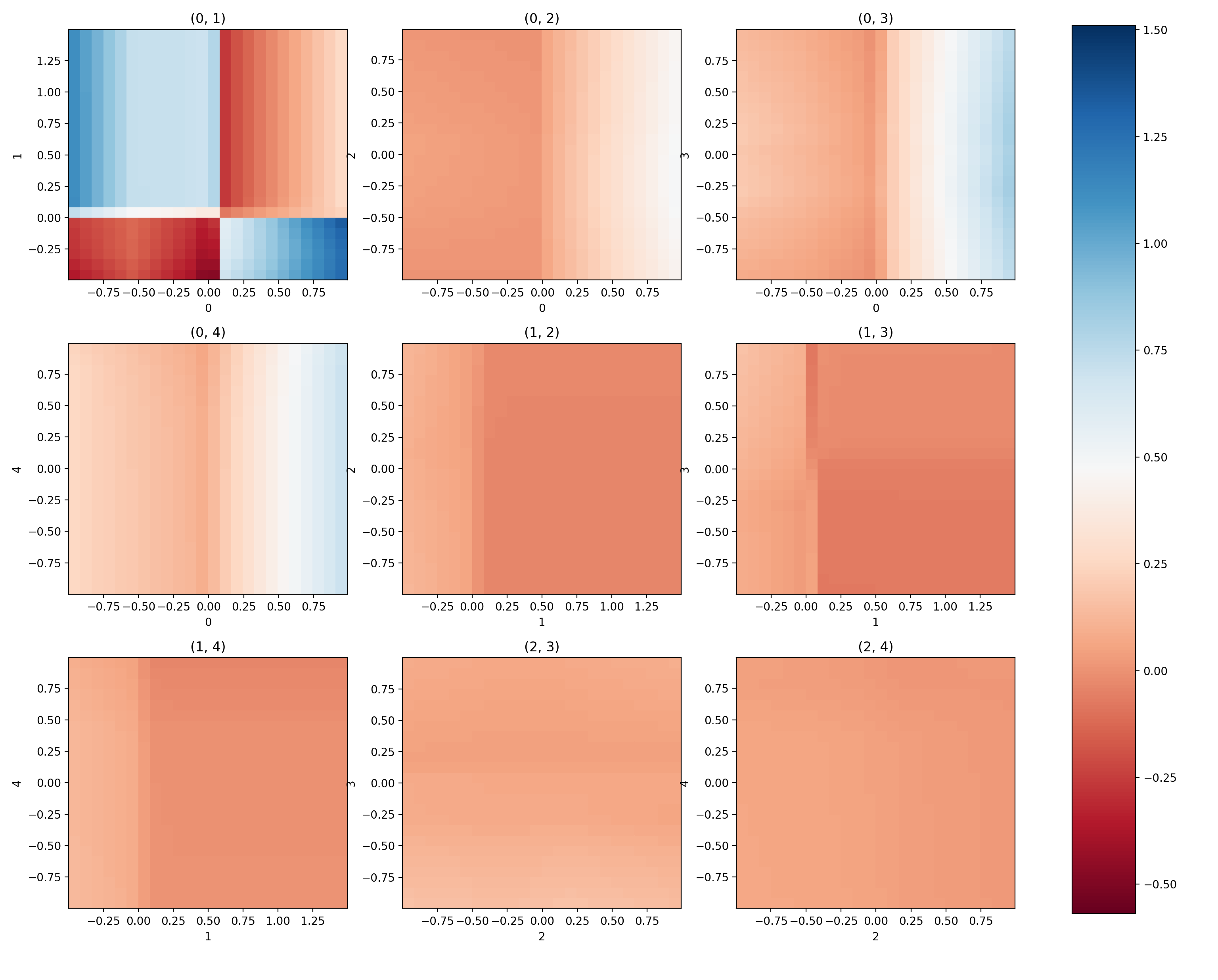}%
\caption{Two-dimensional shape functions in the form of heatmaps illustrating
how pairs of features impact on the target values in the second case of the
chessboard example}%
\label{f:pairwise_shape__chess}%
\end{center}
\end{figure}

Let us consider the second point. For this point, there are dependences of
target values on single features as well as on the pair of the first and
second features. In order to see these pairwise interactions of features, we
consider two-dimensional shape functions in the form of heatmaps illustrating
how pairs of features impact on the target values. The heatmaps for some pairs
of features, which show at least small changes, are depicted in Fig.
\ref{f:pairwise_shape__chess}. Numbers of features in each pair are indicated
above the corresponding heatmap. It can be seen from Fig.
\ref{f:pairwise_shape__chess} that the first heatmap corresponding to shape
function of the pair of features $x_{0}$ and $x_{1}$ (pair $(0,1)$) clearly
illustrates that the proposed method copes with pairwise interactions of
features (see also the right picture in Fig. \ref{f:target_chess}). Other
pairs of features are not important, and this fact is shown in the
corresponding heatmaps whose colors are almost not changed.%

\subsubsection{Pairwise multiplication}

\begin{figure}
[ptb]
\begin{center}
\includegraphics[
height=3.1935in,
width=3.9556in
]%
{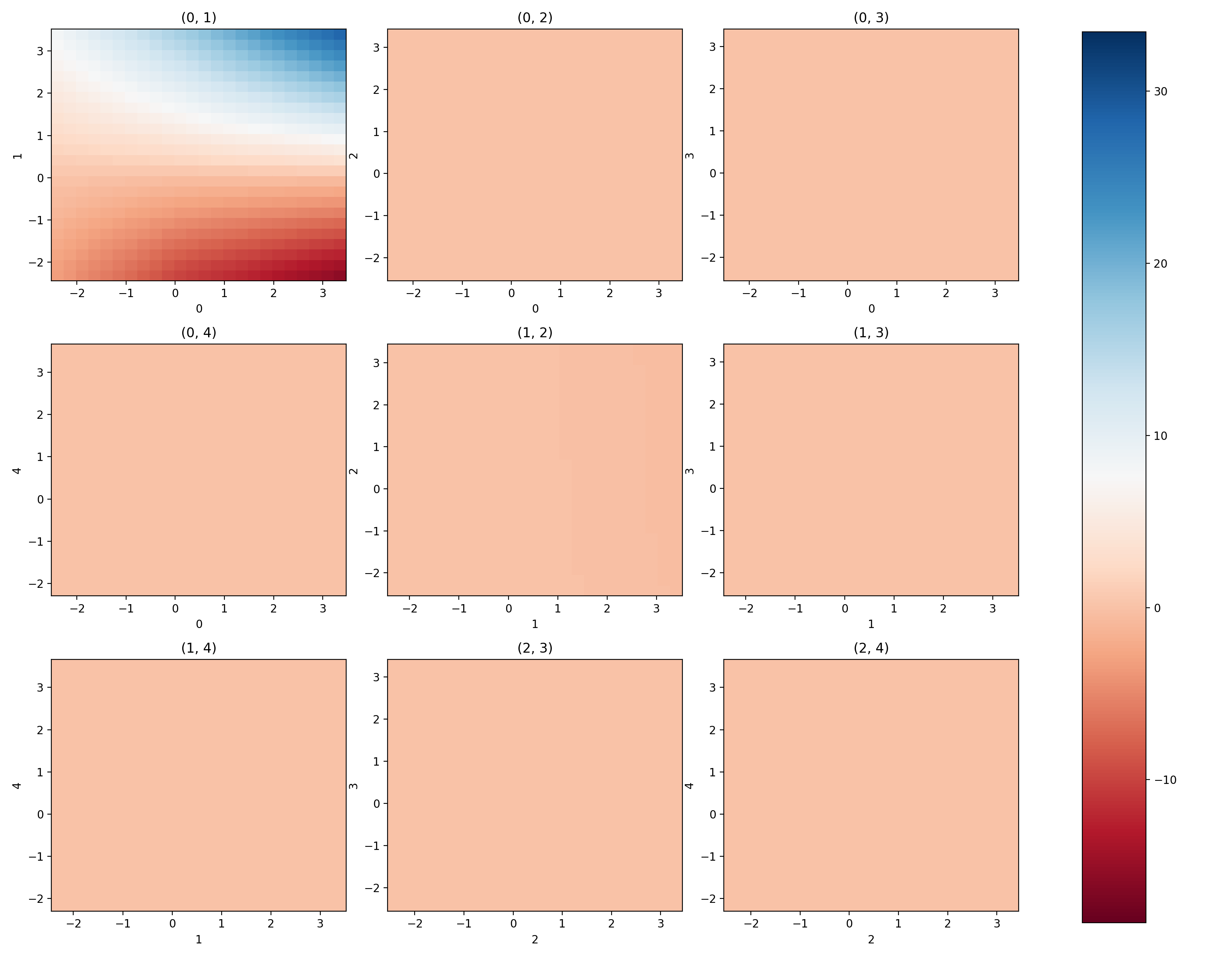}%
\caption{Two-dimensional shape functions in the form of heatmaps illustrating
how pairs of features impact on the target values for the example with
function $x_{0}x_{1}$}%
\label{f:pairwise_mul_shape_1}%
\end{center}
\end{figure}

\begin{figure}
[ptb]
\begin{center}
\includegraphics[
height=3.2259in,
width=3.9898in
]%
{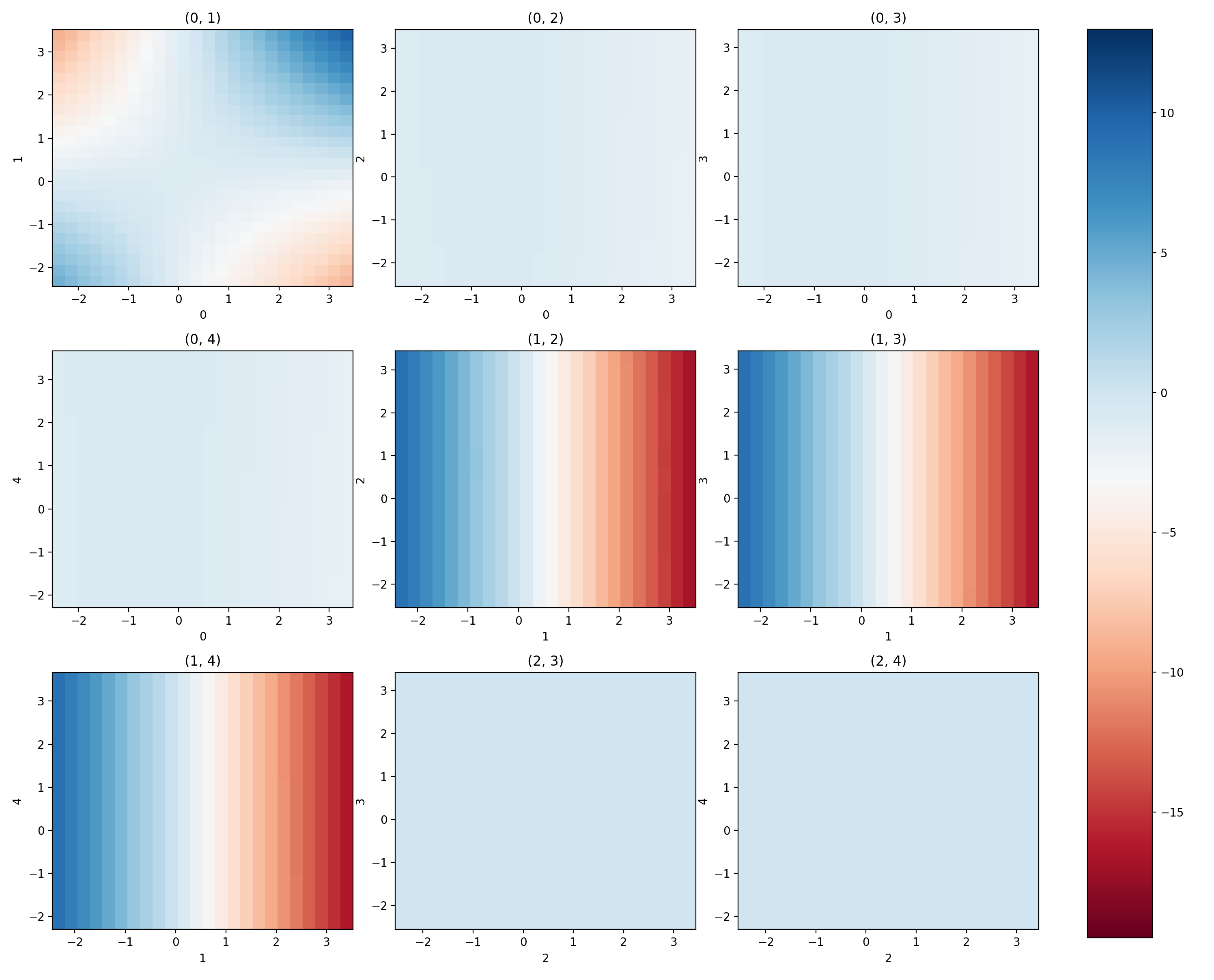}%
\caption{Corrected two-dimensional shape functions in the form of heatmaps
illustrating how pairs of features impact on the target values for the example
with function $x_{0}x_{1}$}%
\label{f:pairwise_mul_shape_2}%
\end{center}
\end{figure}

We again consider a function of five variables, which is a simple product
function $x_{0}x_{1}$. Other variables $x_{2}$, $x_{3}$, $x_{4}$ do not impact
on the target values. Let us take point $(\frac{1}{2},\frac{1}{2},\frac{1}%
{2},\frac{1}{2},\frac{1}{2})$ for explanation the corresponding prediction.
When considering pairwise interactions, it makes sense to rely both on
functions of pairs of features and on functions consisting of the sums of
shape functions of each feature and the function of the pair. In this example,
the black-box model is represented as the product of the first and second
features. Functions depending on pairs of features are close to constant for
all pairs of features, except for the pair $(0,1)$. This fact can be seen from
Fig. \ref{f:pairwise_mul_shape_1} where heatmaps of pairs of features are
depicted similarly to Fig. \ref{f:pairwise_shape__chess}. However, the
dependence itself is not correctly represented because the largest values of
function $x_{0}x_{1}$ should be in the upper right corner and in the lower
left corner. We have to add shape functions of the first and second features.
In this case, we get a correct visualization of the dependence (see Fig.
\ref{f:pairwise_mul_shape_2}). Adjusted shape functions are built as follows:
\begin{equation}
w_{ij}\cdot\mathbf{g}_{i}^{j}+w_{sl}\cdot\mathbf{g}_{s}^{l}+w_{i,s,t}%
\cdot\mathbf{g}_{i}^{j}\cdot\mathbf{g}_{s}^{l}.
\end{equation}
%

%

\subsubsection{Vertical wedge}

\begin{figure}
[ptb]
\begin{center}
\includegraphics[
height=2.1278in,
width=2.0476in
]%
{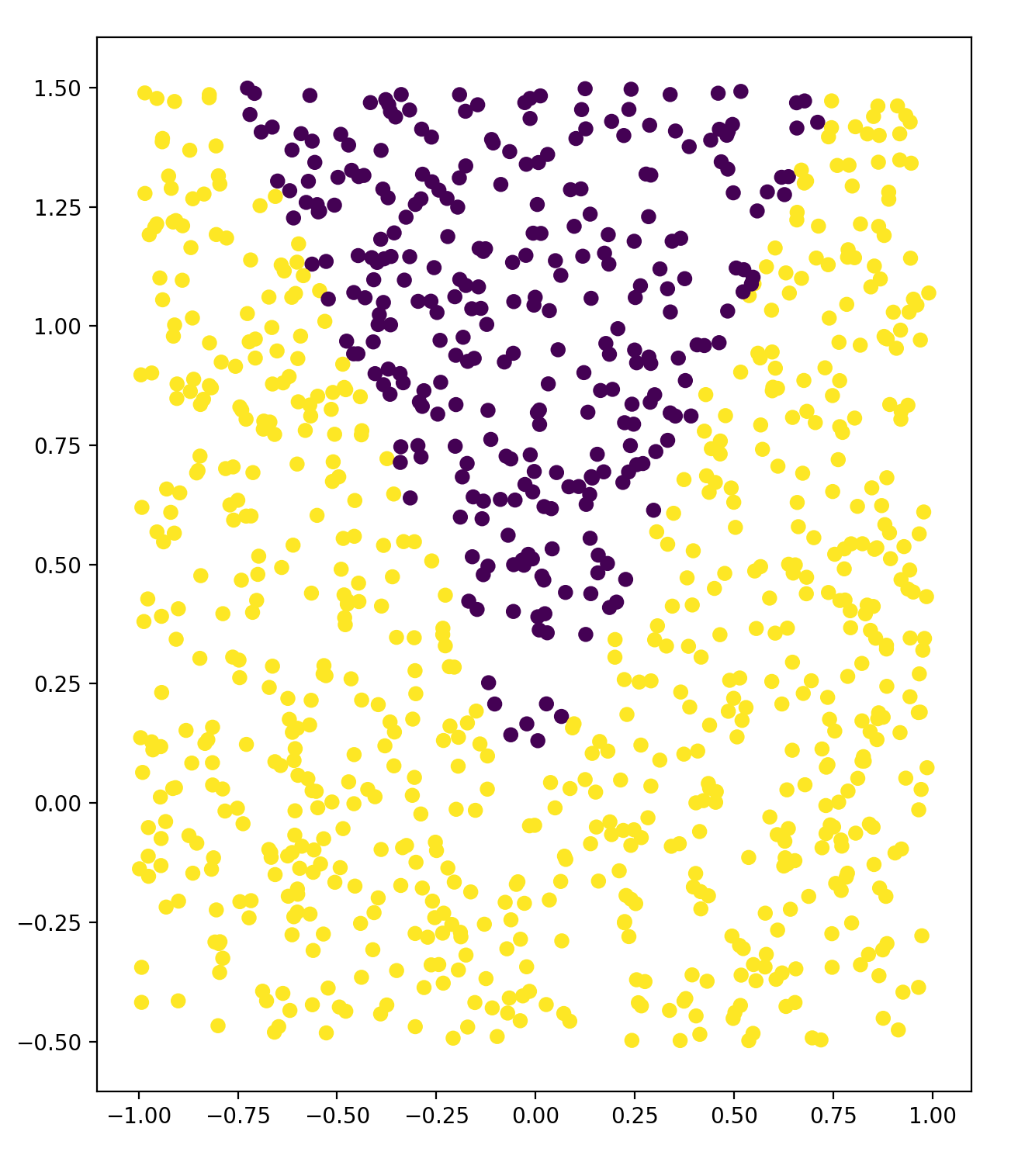}%
\caption{Points of the dataset generated by the \textquotedblleft vertical
wedge\textquotedblright\ function}%
\label{f:target_wedge}%
\end{center}
\end{figure}

\begin{figure}
[ptb]
\begin{center}
\includegraphics[
height=1.3666in,
width=4.9312in
]%
{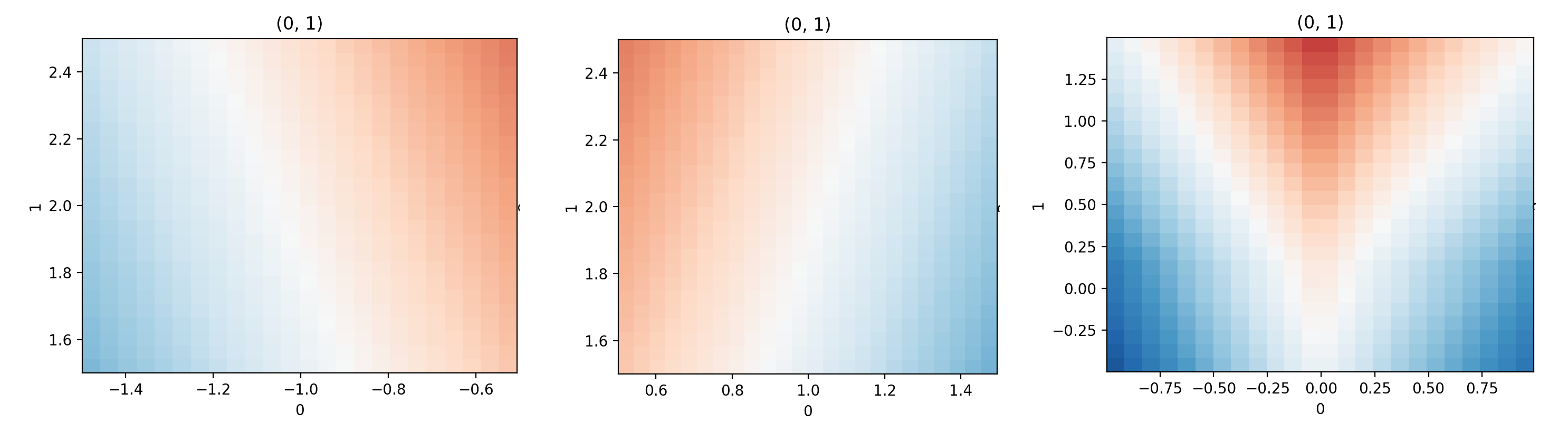}%
\caption{Two-dimensional shape functions in the form of heatmaps illustrating
how pairs of features impact on the target values for the first case of the
\textquotedblleft vertical wedge\textquotedblright\ example corresponding to
three points $(-1,2,0,0,0)$, $(1,2,0,0,0)$, $(0,1/2,0,0,0)$, respectively}%
\label{f:wedge_shape_1}%
\end{center}
\end{figure}

\begin{figure}
[ptb]
\begin{center}
\includegraphics[
height=1.4693in,
width=4.9816in
]%
{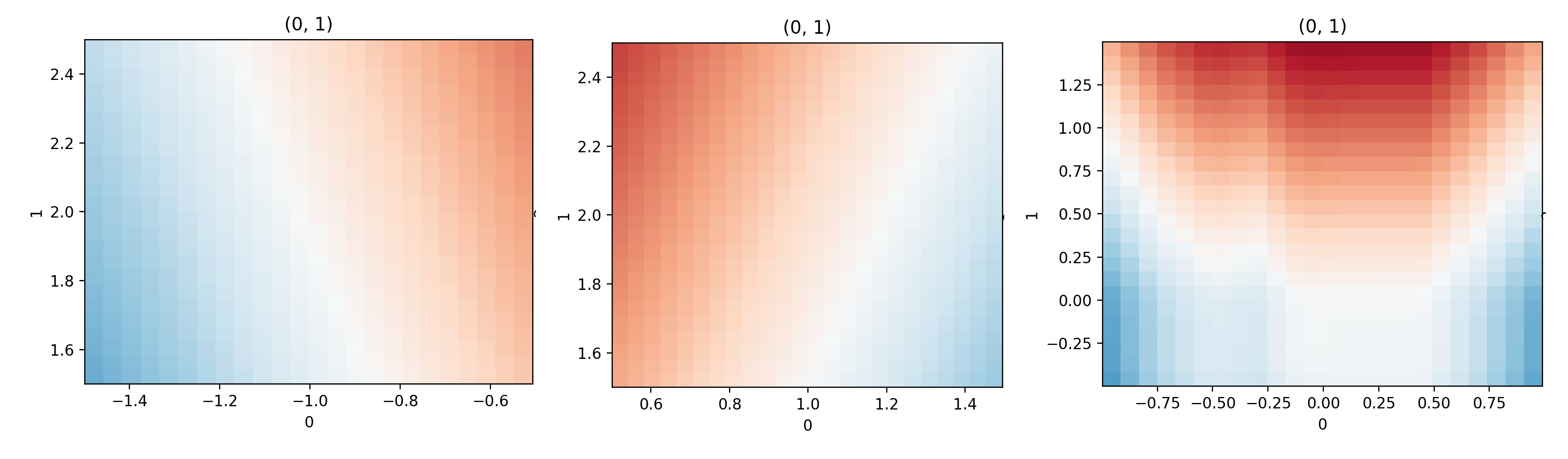}%
\caption{Two-dimensional shape functions in the form of heatmaps illustrating
how pairs of features impact on the target values for the second case of the
\textquotedblleft vertical wedge\textquotedblright\ example corresponding to
three points $(-1,2,0,0,0)$, $(1,2,0,0,0)$, $(0,1/2,0,0,0)$, respectively}%
\label{f:wedge_shape_2}%
\end{center}
\end{figure}

Let us consider a dataset which is generated by the function $I[2\cdot
|x_{0}|>|x_{1}|]$. This function looks like a vertical wedge (see Fig.
\ref{f:target_wedge}). We again consider $5$ features such that only two
features are important. The example can be illustrated by two cases. In the
first case, the black-box model provides predictions in the form of integer
values of classes (a classification problem). In the second case, the
black-box model (the gradient boosting machine consisting of $100$ trees with
largest depth $3$ and learning rate $0.1$) provides probabilities of classes.
Three points are taken for explanation: to the left (feature vector
$(-1,2,0,0,0)$) from the wedge, to the right (feature vector $(1,2,0,0,0)$)
from the wedge and in the center (feature vector $(0,1/2,0,0,0)$). Fig.
\ref{f:wedge_shape_1} depicts heatmaps illustrating how pairs of features
impact on the target values for the first case of the \textquotedblleft
vertical wedge\textquotedblright\ example corresponding to three points
$(-1,2,0,0,0)$, $(1,2,0,0,0)$, $(0,1/2,0,0,0)$, respectively. Only heatmaps
for the pair of features $x_{0}$ and $x_{1}$ are shown because other pairs of
features do not impact on predictions. It can be seen from Fig.
\ref{f:wedge_shape_1} that it is possible to correctly approximate pairwise
interactions of features for all three points. The same heatmaps for the
second case, when probabilities of classes are used as predictions of the
explained black-box model, are depicted in Fig. \ref{f:wedge_shape_2}.%

%

%

\subsection{Real data}

\subsubsection{Boston Housing dataset}

\begin{figure}
[ptb]
\begin{center}
\includegraphics[
height=1.9782in,
width=3.2079in
]%
{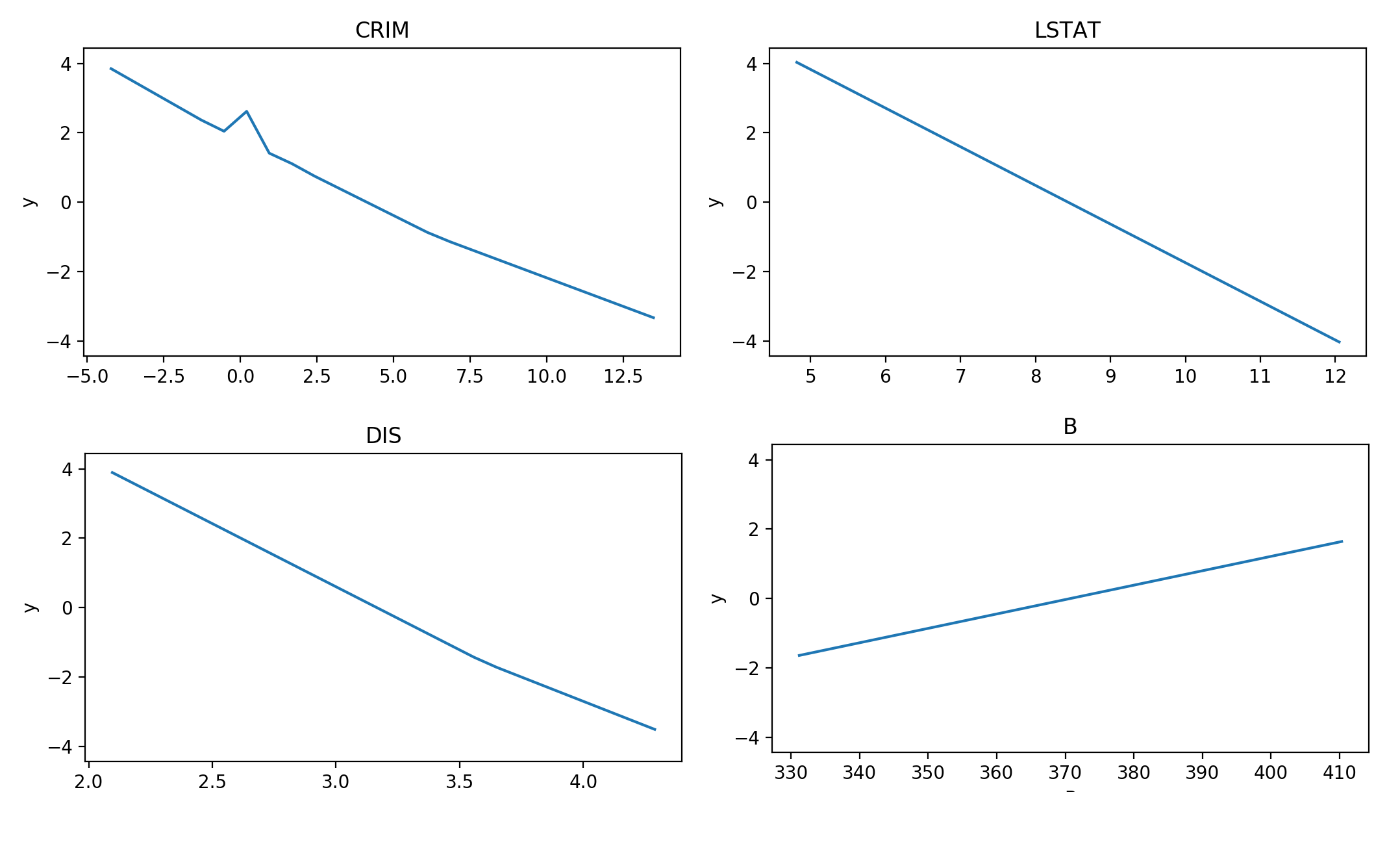}%
\caption{Shape functions of four important features learned on the Boston
Housing dataset for local explanation (the x-axis indicates values of each
feature, the y-axis indicates the feature contribution to the median value of
owner-occupied homes)}%
\label{f:shape_boston_1}%
\end{center}
\end{figure}

Let us consider the real data called the Boston Housing dataset. It can be
obtained from the StatLib archive (http://lib.stat.cmu.edu/datasets/boston).
The Boston Housing dataset contains $506$ examples such that each example is
characterized by $13$ features. The explained black-box model is implemented
as a neural network having 3 layers, every layer consists of 100 units with
the ReLU activation function, and optimizer Adam is used for training on the
Boston Housing dataset. The learning rate is $10^{-3}$, the number of epochs
is $1000$. The proposed system is trained on generated local areas around
randomly selected points from the whole dataset. $N=1000$ perturbed points are
generated around the point of interest $\mathbf{x}$ from the uniform
distribution with bounds $\mathbf{x}_{left}$ and $\mathbf{x}_{right}$ defined
as the 10\%-interval of the feature range around point $\mathbf{x}$ for
explanation, and shape functions are constructed for the local explanation.
Point $\mathbf{x}$ for explanation is taken as the mean of two randomly
selected instances from the Boston Housing dataset. We plot obtained
individual shape functions for four important features \textquotedblleft
CRIM\textquotedblright, \textquotedblleft LSTAT\textquotedblright,
\textquotedblleft DIS\textquotedblright, \textquotedblleft B\textquotedblright%
\ in Fig \ref{f:shape_boston_1}. Similar results have been obtained in
\cite{Konstantinov-Utkin-21}. At that, one can see from the shape plots that
contribution of feature \textquotedblleft B\textquotedblright\ to the median
value of the owner-occupied homes rises while contributions of other features
are decreased.%

\subsubsection{Breast Cancer dataset}

\begin{figure}
[ptb]
\begin{center}
\includegraphics[
height=2.8593in,
width=4.2799in
]%
{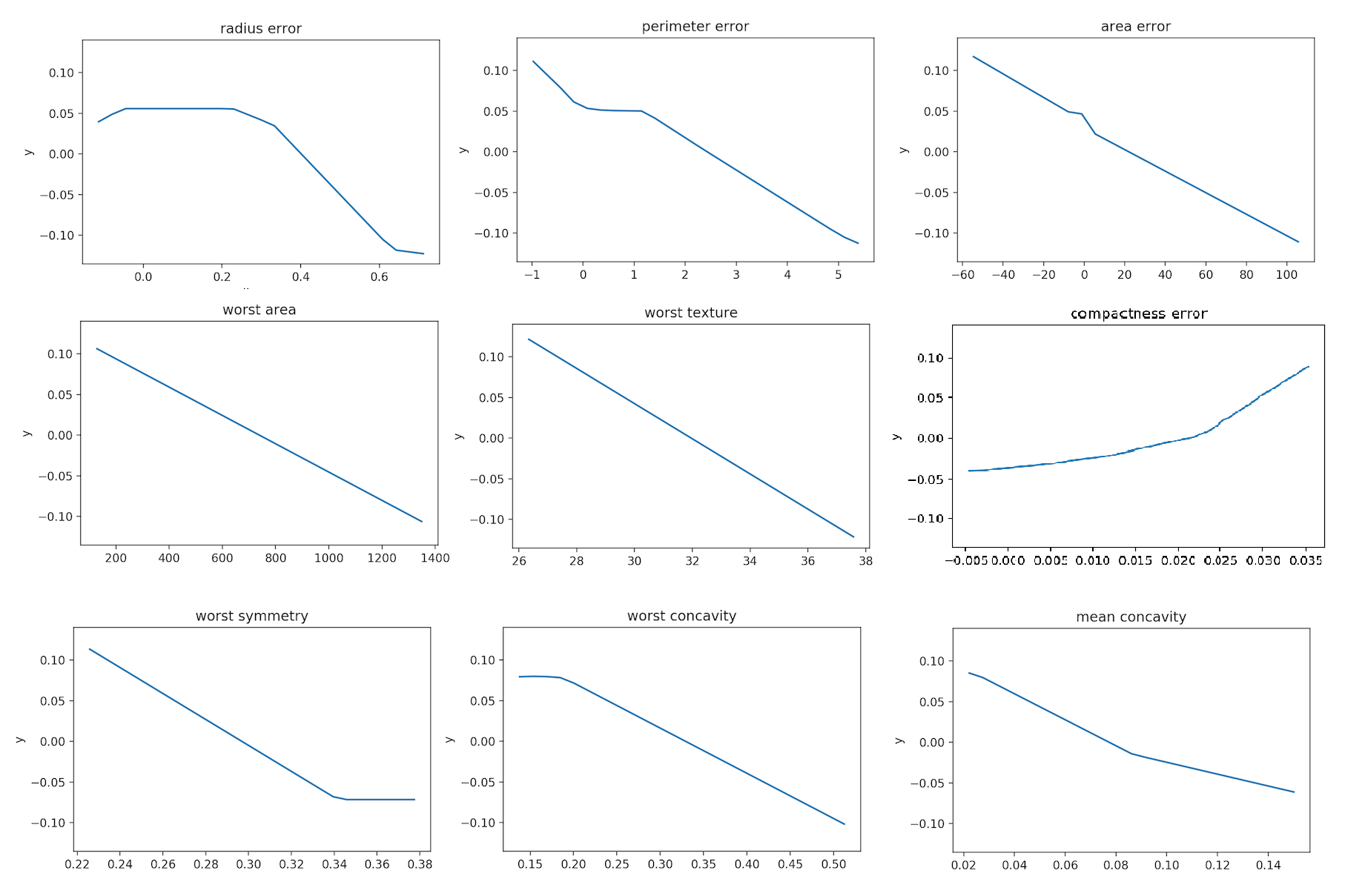}%
\caption{Shape functions of four important features learned on the Breast
Cancer dataset for local explanation (the x-axis indicates values of each
feature, the y-axis indicates the feature contribution to the benign)}%
\label{f:shape_breast_1}%
\end{center}
\end{figure}

The next real dataset is the Breast Cancer Wisconsin (Diagnostic). It can be
found in the well-known UCI Machine Learning Repository
(https://archive.ics.uci.edu). The Breast Cancer dataset contains $569$
examples such that each example is described by $30$ features. The malignant
and the benign are assigned by classes $0$ and $1$, respectively. We use the
classification black-box model based on the SVM with the RBF kernel having
parameter $\gamma=1/m$ and the penalty parameter $C=0.1$. We plot obtained
individual shape functions for nine important features\ in Fig
\ref{f:shape_breast_1}. Other features are not important because their shape
functions do not change or their changes are very small in comparison with
features whose shape functions are shown in Fig. \ref{f:shape_breast_1}. It is
interesting to point out that features of the highest importance obtained in
\cite{Konstantinov-Utkin-21} for the same random point on the bases of the
Breast Cancer dataset are \textquotedblleft radius error\textquotedblright,
\textquotedblleft area error\textquotedblright, \textquotedblleft worst
area\textquotedblright, \textquotedblleft worst symmetry\textquotedblright.
All these features are included into the set of nine important features shown
in Fig. \ref{f:shape_breast_1}.%

\subsubsection{California Housing dataset}

\begin{figure}
[ptb]
\begin{center}
\includegraphics[
height=2.8466in,
width=2.8737in
]%
{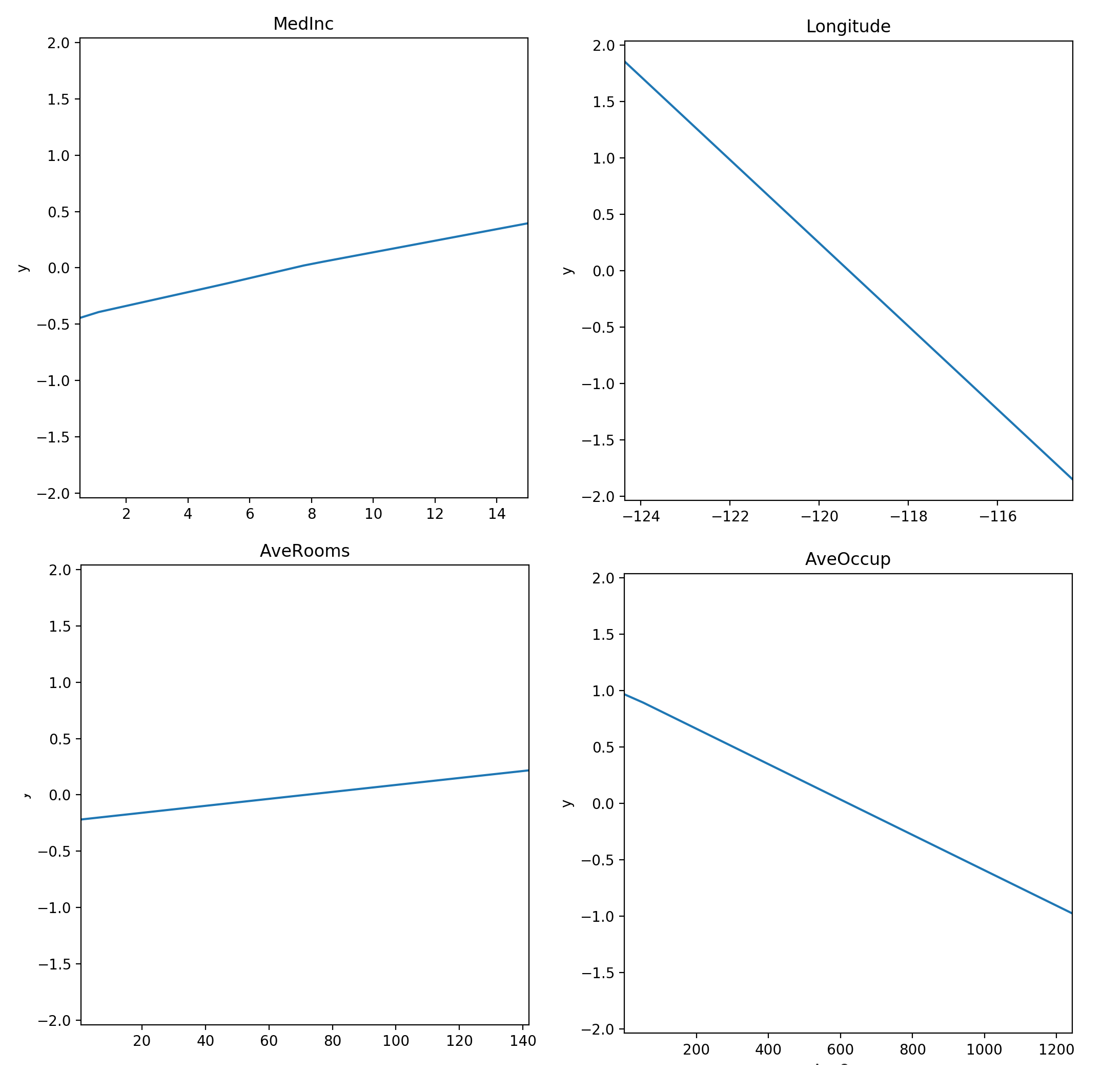}%
\caption{Shape functions of four important features learned on the California
Housing dataset (the x-axis indicates values of each feature, the y-axis
indicates the feature contribution to the median price of houses)}%
\label{f:shape_california_1}%
\end{center}
\end{figure}

Another numerical example is for comparing AFEX with NAM
\cite{Agarwal-etal-20}. This is the California Housing dataset which can be
found in the Kaggle
(https://www.kaggle.com/camnugent/california-housing-prices). It contains
$20640$ observations with $9$ features and aims to solve a regression problem
to predict the median price of houses depending on several parameters. The
explained black-box model is implemented as a gradient boosting machine with
decision trees of depth $3$. It is trained by using $100$ iterations with the
learning rate $0.1$. A point for explanation is taken as a center of the
dataset. It has the following values of features: \textquotedblleft
MedInc\textquotedblright=$3.87$; \textquotedblleft HouseAge\textquotedblright%
=$28.64$; \textquotedblleft AveRooms\textquotedblright=$5.43$;
\textquotedblleft AveBedrms\textquotedblright=$1.1$; \textquotedblleft
Population\textquotedblright=$1425.5$; \textquotedblleft
AveOccup\textquotedblright=$3.1$; \textquotedblleft Latitude\textquotedblright%
=$35.6$; \textquotedblleft Longitude\textquotedblright=$-119.6$. The global
explanation as in \cite{Agarwal-etal-20} is studied, therefore, an area around
the point is defined by lowest and largest values of all features. Shape
functions of four important features \textquotedblleft
MedInc\textquotedblright, \textquotedblleft Longitude\textquotedblright,
\textquotedblleft AveRooms\textquotedblright, \textquotedblleft
AveOccup\textquotedblright\ are depicted in Fig. \ref{f:shape_california_1}.
Other features cannot be viewed as important. The same results have been
obtained by means of NAM \cite{Agarwal-etal-20}. However, the authors of
\cite{Agarwal-etal-20} pointed out that only two features \textquotedblleft
MedInc\textquotedblright\ and \textquotedblleft Longitude\textquotedblright%
\ should be regarded as important.%

\begin{figure}
[ptb]
\begin{center}
\includegraphics[
height=2.8638in,
width=4.7366in
]%
{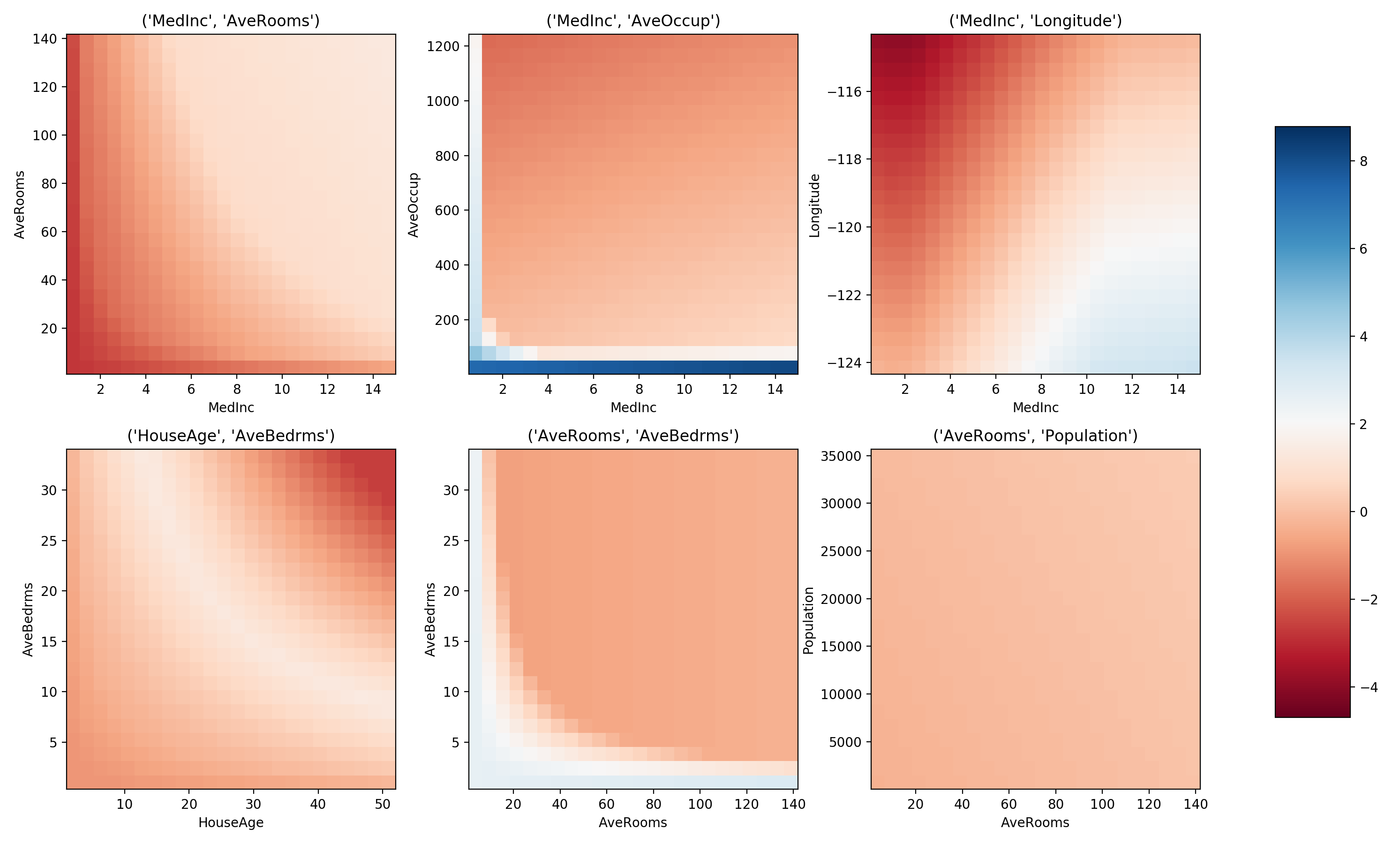}%
\caption{Two-dimensional shape functions in the form of heatmaps illustrating
how pairs of features impact on the target values for California Housing
dataset}%
\label{f:pairwise_california}%
\end{center}
\end{figure}

In order to study the pairwise feature interactions, we also construct
heatmaps for pairs of features which can be viewed as important. They are
depicted in Fig. \ref{f:pairwise_california}. It is surprisingly seen from
Fig. \ref{f:pairwise_california} that the pair of features \textquotedblleft
MedInc\textquotedblright\ and \textquotedblleft Longitude\textquotedblright%
\ is important, but the pair \textquotedblleft MedInc\textquotedblright\ and
\textquotedblleft AveOccup\textquotedblright\ is more important because small
values of feature \textquotedblleft AveOccup\textquotedblright\ lead to the
crucial change of the feature contribution.%

\section{Conclusion}

A new approach for solving the explanation problem of machine learning
black-box model predictions has been proposed. System AFEX implementing the
approach has several important advantages. First, AFEX\ identifies pairwise
interactions between features in a simple way. Second, the system for a
dataset is trained once and the local explanation of arbitrary points in the
area of the dataset does not require to train again. Numerical experiments
with synthetic and real data have demonstrated that AFEX correctly explain
various difficult cases and datasets.

On the one hand, we have pointed out that shape functions may be a better way
for representing results especially when pairwise interactions are studied. On
the other hand, explanations are simpler for many users when they are
represented by a single number measuring the feature importance. Therefore,
development of approaches to represent shape functions, especially heatmaps,
in a form of point-valued feature contributions is an interesting task for
further research.

One of the ideas which has been proposed in AFEX is to multiply all pairs of
columns corresponding to different features in order to realize the method for
the pairwise interaction study. At the same time, the product of two vectors
is one of the possible operations for taking into account the pairwise
interactions. The flexibility of AFEX allows us to consider other operations
different from the standard product. Study of these operations and
investigation how they impact on the explanation results can be regarded as
another direction for further research.

Another problem of AFEX is that the number of neural subnetworks is equal to
the number of features multiplied by the value of the basis shape functions
$k$. Hence, high-dimensional datasets cannot be considered due to difficulties
of training a huge amount of subnetworks. One of the ways to cope with this
difficulty is to use a machine learning model different from neural network,
for example, the gradient boosting machine. A justified selection of the best
model for feature transformation is also a direction for further research and study.

We have applied only one differentiable model for computing weights, namely,
the linear regression model. However, it is interesting to investigate whether
AFEX can be improved by using another differentiable model, for example, SVM.
This is another direction for research in future.

AFEX is developed for dealing with tabular data. Its modification for
processing images, graph data, text information is also an interesting task
for further research.

\section*{Declaration of competing interest}

The authors declare that they have no known competing financial interests or
personal relationships that could have appeared to influence the work reported
in this paper.

\section*{Acknowledgement}

This work is supported by the Russian Science Foundation under grant 21-11-00116.

\section*{Acknowledgement}

The research is partially funded by the Ministry of Science and Higher
Education of the Russian Federation as part of World-class Research Center
program: Advanced Digital Technologies (contract No. 075-15-2020-934 dated 17.11.2020)

\bibliographystyle{unsrt}
\bibliography{Boosting,Expl_Attention,Explain,Explain_med,Lasso,MYBIB,Robots}

\end{document}